\documentclass[10pt,twocolumn,letterpaper]{article}

\usepackage{cvpr}
\usepackage{times}
\usepackage{epsfig}
\usepackage{graphicx}
\usepackage{amsmath}
\usepackage{amssymb}
\usepackage{algorithm}    
\usepackage{algorithmic}
\usepackage{subfigure}
\usepackage{multirow}
\usepackage{float}
\usepackage{color}

\usepackage[breaklinks=true,bookmarks=false]{hyperref}

\cvprfinalcopy 

\def\cvprPaperID{2961} 

\ifcvprfinal\fi
\begin{document}

\title{Adaptive Re-ranking of Deep Features for Person Re-identification}

\author{Yong Liu\\
Nanjing University\\
{\tt\small mg1733041@smail.nju.edu.cn }
\and
Lin Shang\\
Nanjing University\\
{\tt\small shanglin@nju.edu.cn}
\and
Andy Song\\
RMIT University\\
{\tt\small andy.song@rmit.edu.au}
}

\maketitle

\begin{abstract}
State-of-the-art person re-identification (re-ID) methods train a deep CNN to extract features and combine these features with a distance metric for final evaluation. In this work, we aim to improve re-ID performance by better utilizing the deeply-learned features. First, we propose an Adaptive Re-Ranking (ARR) method to better incorporate the contextual information embedded in the neighboring features. ARR re-ranks the match results in an iterative manner based on contextual neighborhood structure. In particular, it automatically adds more contextual information after each iteration in order to find more matches. Second, we propose a Deep Feature Fusion (DFF) method to exploit the diverse information in the deeply-learned features. DFF splits these features into sub-features and exchanges feature information by a diffusion process.  Consequently, we propose a strategy that combines ARR and DFF to further enhance re-ID performance. Extensive comparative evaluations on three large re-ID benchmarks show that our method is robust and highly competitive.
\end{abstract}

\section{Introduction}
\begin{figure}[!t]
\centering
\includegraphics[width=8cm,height=3.4cm]{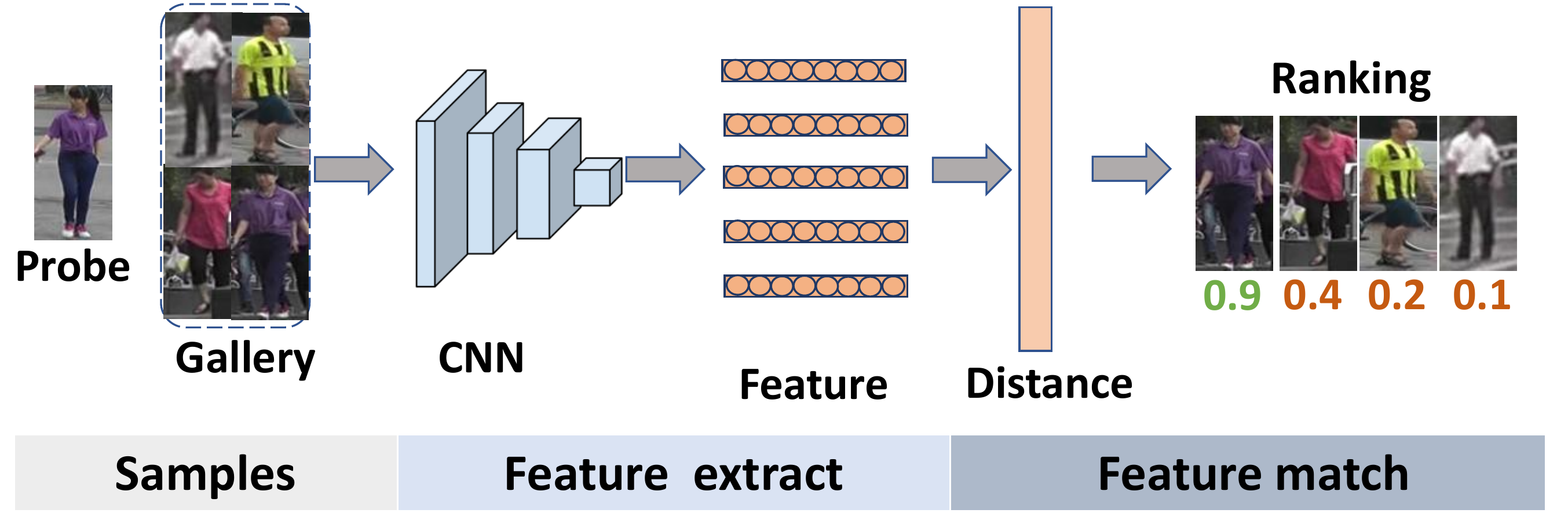}  
\caption{The pipeline of person re-identification.}
\label{pipeline}
\end{figure}

Person re-identification (re-ID) \cite{gong2014person} aims to recognize and associate a person of interest across non-overlapping cameras. It becomes increasingly more important but still remains a challenge as the same person observed in various cameras may appear different due to the large intra-class variations in illumination, poses, occlusions, and backgrounds. To compensate the appearance variation across cameras, most previous studies focus on two aspects: (1) extracting effective and robust visual features \cite{Liao2015Person,Matsukawa2016Hierarchical,Yang2014Salient}; (2) learning an effective metric \cite{Hirzer2012Relaxed,Xiong2014Person,yan2014view}. With the success of deep learning, deeply-learned features (deep features for short) have demonstrated superior performance against hand-crafted ones and become dominant in the field of re-ID. A typical pipeline of deep feature based re-ID consists of two steps: feature extract and feature match. It first utilizes a part of the well-trained network as a feature extractor and then uses a certain distance metric such as Euclidean distance for person retrieval (see Fig.~\ref{pipeline}). Images with high enough score are considered as positives.

Re-ranking is an effective step to boost the performance of re-ID \cite{Garcia2015Person,Shen2012Object,Yu2017Divide}, in which the contextual re-ranking is popular in recent years. The intuition of contextual re-ranking is that images containing the same person should not only have similar visual content (\emph{content similarity}), but also possess similar $k$-nearest neighbors (\emph{context similarity}) \cite{Leng2015Person}. The contextual re-ranking takes a distance matrix as input and outputs a better matrix as a new metric using additional contextual information. Once the new metric is obtained, the re-ranking method can be re-applied naturally in an iterative manner. However, existing iterative re-ranking methods in re-ID ignore the fact that the effectiveness of the new metric increases along iterations \cite{Yu2017Divide}. To tackle this problem, we propose an Adaptive Re-Ranking method named ARR, in which the contextual information is increasing along iterations automatically to consider more matches. 

Although combining deep features with Euclidean distance produces promising performance, there is still information that is not fully utilized such as the diversity of different parts of the feature. Inspired by the idea of ``Divide and Fuse'' in \cite{Yu2017Divide}, thereby we propose a method named Deep Feature Fusion (DFF) to incorporate the diversity information involved in the deep feature with a diffusion process which is widely used in object retrieval. 

Both ARR and DFF make certain transformations to the distance metric. They can naturally be incorporated together for further improvements. For example, ARR can take the distance metrics from sub-features as input and outputs new metrics. Multiple new metrics can be fused into a single one by DFF. The details are provided in the following sections. The main contributions of this study are in summary: 
\begin{itemize}
\item An Adaptive Re-Ranking (ARR) method. It utilizes more contextual information automatically after each iteration to consider more matches and can fully exploit the contextual information encoded in neighborhood structures. 

\item A Deep Feature Fusion (DFF) method. It integrates the diversity information involved in sub-features. DFF improves the performance considerably, demonstrating itself as an effective method for re-ID. 

\item A strategy to incorporate ARR and DFF to further improve re-ID performance. 
\end{itemize}

\section{Related Work}
Re-ranking and features fusion are essentially the two areas of focus of this study.  Hence prior works relevant to these two areas are reviewed in this section.

{\bfseries Re-ranking.}
Basically, there are two main streams for re-ranking: manifold ranking and contextual re-ranking. In manifold ranking \cite{Jegou2011A}, a similarity matrix or affinity matrix is interpreted as a graph where the weight of each edge represents the affinity of two data points. A key component of the manifold ranking is the diffusion process. It propagates the affinity information of all data points on the weighted graph. Compared to the traditional pairwise affinity which measures the similarity between two data points, the diffusion process explores the affinity information among all data points in a global view. In addition, the manifold ranking can be ranged into two principal types, diffusion on a single graph \cite{Jegou2011A,yang2009locally,yang2013affinity} and on multiple graphs \cite{luo2013shape,wang2012unsupervised,yang2015re,zhang2015query}. The latter is designed for features fusion and metrics fusion, named as fusion by diffusion. In particular, each affinity matrix (based on a specific feature or metric) is interpreted as a weighted graph and the affinity information is spread on multiple graphs. Fusion by diffusion not only captures the intrinsic manifold structure, but also leverages the complementarity of multiple visual features or multiple metrics. 

In recent years, the contextual re-ranking methods show excellent performance \cite{bai2016sparse,sarfraz2017pose,Yu2017Divide,Zhong2017Re}. In contextual re-ranking, the similarity of two data points is re-defined by the similarity of their rank lists for the same query. The rank list is usually a short list containing the top $k$ nearest neighbors instead of all data points. The motivation for contextual re-ranking is that similar data possess similar neighbors, thus the contextual information is exploited. Contributions on contextual re-ranking are devoted to solving the efficiency problem \cite{bai2016sparse} and effectiveness problem \cite{Yu2017Divide,Zhong2017Re}, or both \cite{sarfraz2017pose}. Motivated by \cite{Yu2017Divide}, we introduce an iterative scheme to the $k$-reciprocal encoding re-ranking \cite{Zhong2017Re} to improve the robustness of re-ranking, named Adaptive Re-Ranking (ARR). Different from \cite{Yu2017Divide}, the contextual information in ARR increases during iterations rather than remain unchanged.

{\bfseries Features Fusion.} Features fusion has been proven effective in object retrieval \cite{rahman2011learning,yu2013feature,zhang2012query,zhang2015query} and re-ID \cite{mirmahboub2018person,zheng2015query}. For example, \cite{zheng2015query} fuses the hand-crafted features and deep features based on an observation: the score curve for a query exhibits an ``L'' shape for a good feature but descends gradually for a bad one. Obtaining multiple features can be burdensome. Therefore, \cite{Yu2017Divide} proposed ``Divide and Fuse'' (DaF) method to exploit the information involved in a deep feature under the framework of features fusion. The basis of \cite{Yu2017Divide} is that the characteristics of different parts of a deep feature (typically with a high dimension) can be rather diverse \cite{Yu2017Divide}. DaF divides a feature into parts, namely sub-features, instead of treating a feature vector as a whole. DaF then fuses sub-features by a fuzzy aggregation operator in fuzzy theory. 

The fusion strategies for multiple features \cite{mirmahboub2018person,zheng2015query} or multiple sub-features \cite{Yu2017Divide} in the existing works can be classified as the weighted linear combination. But they did not utilize the rich information embedded in the relationships among data points \cite{pedronette2011exploiting}. The fusion by diffusion process reviewed in this section can address this issue effectively. In this paper, we propose a Deep Feature Fusion (DFF) method which simultaneously leverages ``Divide and Fuse'' and the fusion by diffusion process. In particular, DFF splits the deep feature into multiple sub-features evenly and then fuses them for re-ID.

\section{Adaptive Re-ranking of Deep Features}
\begin{figure*}[!t]
\centering
\includegraphics[width=12cm,height=5.4cm]{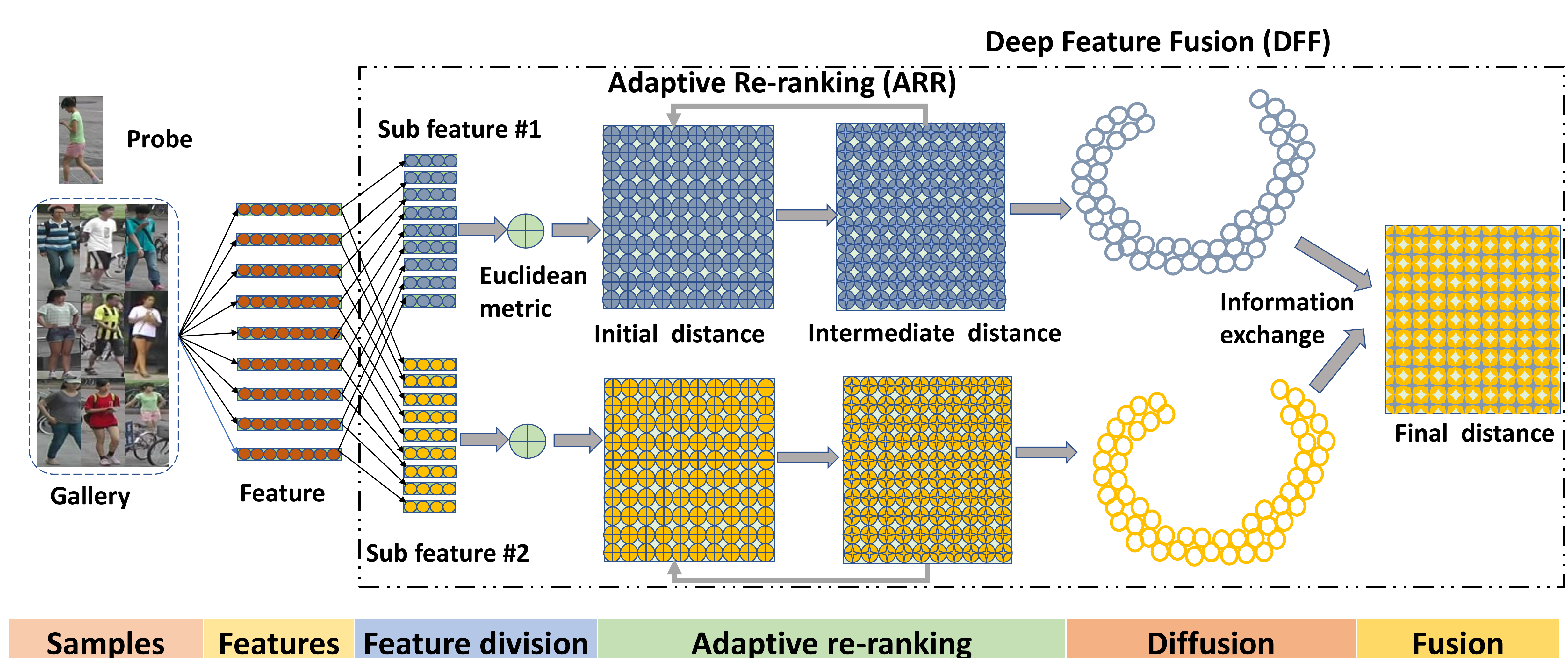}  
\caption{Overview of the proposed method for person re-identification in the case of $S$ ($S=2$) sub-features.}
\label{all}
\end{figure*}
The overall architecture of the proposed approach is illustrated in Fig.~\ref{all}. Firstly, each deeply trained feature is evenly split into $S$ sub-features. So $S$ distance metrics are obtained. Secondly, the adaptive re-ranking method re-defines each metric to a new metric. Thirdly, $S$ re-defined metrics are fused into a single one by the deep feature fusion for the final evaluation. The details of adaptive re-ranking and deep feature fusion are presented in Section 3.1 and 3.2 respectively. In Section 3.3, we describe how to combine ARR and DFF. 

\subsection{Adaptive Re-ranking (ARR)}
We first revisit the conventional contextual re-ranking. Without loss of generality, we consider a probe $p$ from camera $a$ and a gallery set $G = \{g_i| i=1,2,\cdots,M\}$ with $M$ images from camera $b$. The Euclidean distance for an image pair $(p,g_i)$ is calculated as:
 \begin{equation}
 D_E(p,g_i)=||\mathbf p-\mathbf g_i||_2^2. 
   \end{equation}
 The goal of re-ID is to search in $G$ for images containing the same person with $p$. Therefore, $G$ is sorted in an ascending order according to their distance to $p$ and the initial rank is obtained:
 \begin{equation}
 L(p,G)=\{g_{(i)}| i=1,2,\cdots,M\}
  \end{equation}
 The subscript $(i)$ is enclosed within parenthesis indicating that $i$ is the position in $L(p,G)$ instead of $G$. Images in the top ranks of $L(p,G)$ are likely to be the true matches. Most of the possible false matches can be removed easily with $L(p,G)$. However, some false matches may also be included in the top ranks. Hence, re-ranking is an important step to boost the performance. After re-ranking, the ranks of true matches will rise and that of false alarms will decline.

 The goal of re-ranking is to learn a function:
  \begin{equation}
f:D \rightarrow D'
  \end{equation}
It takes a distance metric $D$ (usually Euclidean distance) as input and outputs a more effective one. With the new metric, a more faithful rank is obtained. The core idea of contextual re-ranking is that the distance between two points is replaced by the generalized distance between their $k$-nearest neighbors. But the $k$-nearest neighbor relationship is asymmetric since $q_2 \in top_k(q_1) \nRightarrow q_1 \in top_k(q_2)$, where $top_k(q)$ represents $k$-nearest neighbors of $q$. To encode stronger neighborhood information, \cite{qin2011hello} defines the $k$-{reciprocal} neighbor relationship as follows:
\begin{equation}
 r_k(q_1,q_2)=q_2 \in top_k(q_1)  \&  q_1 \in top_k(q_2)
 \end{equation}

$r_k(q_1,q_2)$ is either 0 or 1. $r_k(q_1,q_2)=1$ indicates that $q_1$ and $q_2$ are the $k$-nearest neighbors with each other. The $k$-reciprocal neighbor relationship is obviously symmetric and encodes stronger neighborhood information than the $k$-nearest neighbor relationship. Based on the $k$-reciprocal neighbor relationship, the $k$-reciprocal feature of the image $q_1$ is computed as:
\begin{equation}
R(q_1)=\{q_2 |r_k(q_1,q_2)=1 \}
  \end{equation}
Note that $R(*)$ is a set instead of a vector. Correspondingly, the distance of a pair $(q_1,q_2)$ is measured by the generalized Jaccard distance:
\begin{equation} 
D_J(q_1, q_2)=1-\frac{|R(q_1) \cap R(q_2)|}{|R(q_1) \cup R(q_2)|}
\label{jaccard}
\end{equation}
The smaller $d_J$ is, the more co-neighbors between $q_1$ and $q_2$. Except for some changes in detail, most of the re-ranking methods in re-ID follow a similar procedure as described above. The $k$-reciprocal encoding re-ranking \cite{Zhong2017Re} is adopted in this paper due to its simplicity and effectiveness. It incorporates $D_E$ and $D_J$ to obtain a more robust metric, where $D_E$ represents Euclidean distance. The final distance of an image pair ($q_1, q_2$) is calculated as:
\begin{equation} 
D_F(q_1, q_2)= D_E( q_1, q_2)+ D_J( q_1, q_2)
\label{finaldist}
\end{equation}
\begin{figure}[b!]
\centering
\includegraphics[width=8.5cm]{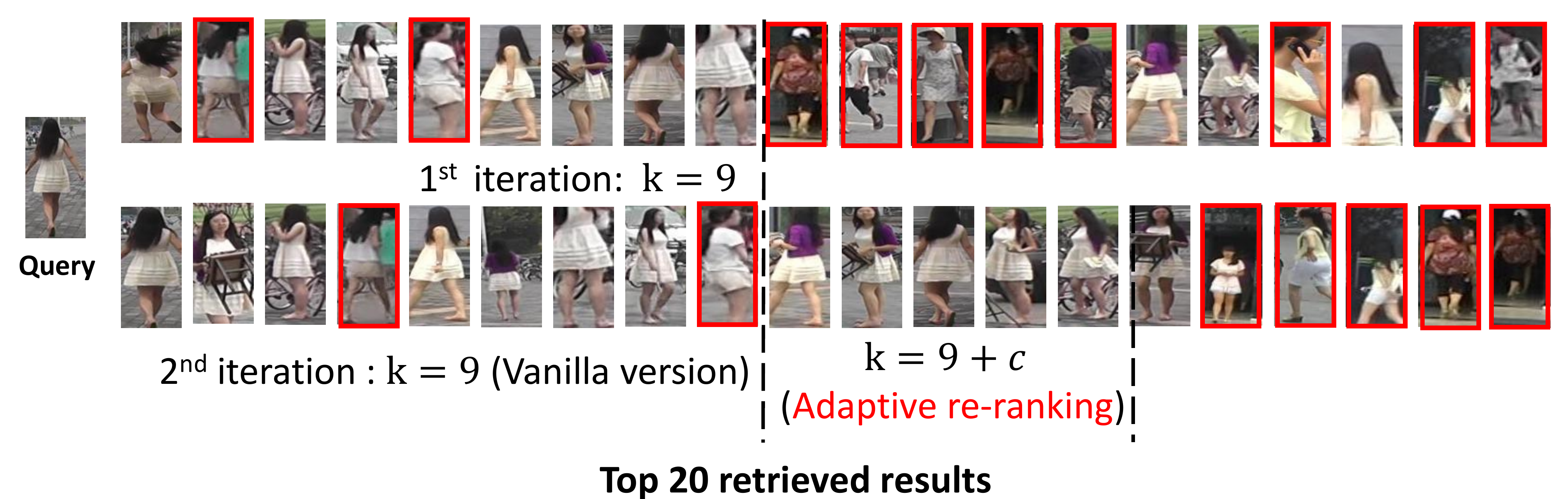} 
\caption{Example of the adaptive re-ranking on the Market-1501 \cite{zheng2015scalable} dataset. Images enclosed by red boxes are false matches. The effectiveness of the distance metric is increasing along iterations. }
\label{whyKIncrease}
\end{figure}
Recall that the input and output of re-ranking algorithms are both distance metrics. An iterative strategy can be adopted to it naturally. However, the existing iterative re-ranking algorithm \cite{Yu2017Divide} in re-ID ignores the fact that the effectiveness of the distance metric is increasing along iterations.

 Fig.~\ref{whyKIncrease} shows an example of top-20 ranks for a given query along two iterations. After the first round of re-ranking, there are 10 false matches and most of them receive high ranks. After the second round, 7 false matches with relatively lower ranks are included in top-20 ranks. Existing iterative re-ranking method, refers to \emph{vanilla version}, sets the number of nearest neighbors to a fixed value, for example, $k$=9. In fact, $k$ implies the amount of contextual information that can be utilized in the contextual re-ranking. The larger the $k$, the more contextual information is exploited by the contextual re-ranking. But excessive contextual information may contain some noise information and degrades the performance. Therefore, $k$ is usually set to a small value conservatively as shown in the first round of Fig.~\ref{whyKIncrease}. However, in an iterative re-ranking method, the effectiveness of the distance metric increases along iterations so more contextual information can be utilized to boost the performance. Therefore, the adaptive re-ranking method ARR is designed. 

 As shown in Fig.~\ref{whyKIncrease}, $k$ is increased to $k+c$ after each iteration, where $c$ is a constant. By this approach more matching images can be considered. 

\subsection{Deep Feature Fusion (DFF)}
In typical implementations of re-ID, deep features are extracted from deep neural networks for distance measures. The Euclidean distance is used as the metric for the final evaluation. Inspired by \cite{Yu2017Divide}, we split a good deep feature vector evenly into multiple sub-features and then fuse them. 

This method is more suitable for deep features than hand-crafted ones. In our preliminary experiments, fusing multiple hand-crafted sub-features sometimes impairs the performance. As pointed out in \cite{Yu2017Divide}, deep features extracted from the fully connected layer are unbiased while hand-crafted ones often maintain a certain structure. Therefore, we believe that splitting the latter into sub-features may undermine their structural information.

Formally, given person images from $C$ non-overlapping cameras, a deep feature vector is extracted from each image and split into $S$ sub-features. The sub-features produce different distance metrics, denoted as $\tilde D=\{D_s|s=1,2,\cdots,S\}$. Specifically, there are two stages: graph construction and multi-graph fusion. We first construct a graph for every metric. Each metric $D_s$ is interpreted as an affinity matrix of a finite weighted graph $G^s=(V_s,W_s)$, where the vertex set $V_s$ represents all images. $V_s$ is connected by edges with edge strength $W_s$. Each element $W_s(i,j)$ in $W_s$, representing the similarity of image pair ($i$,$j$), is calculated as:
\begin{equation}
W_s(i,j)=exp(-D_s(i,j))
\end{equation}
A Markov chain is constructed for each graph with the transition probability defined as:
\begin{equation}
P_s(i,j)=\frac {W_s(i,j)}{\sum_{k=1}^N W_s(i,k)}
\end{equation}
where $\sum_{k=1}^N W_s(i,k)$ is the degree of the vertex. $P_s(i,j)$ represents the transition probability of $i$ to $j$ under the $s$-th sub-feature. The more similar $i$ is to $j$, the more likely that it will jump to $j$. To further reduce the influence of noise points, a local transition probability is used:
\begin{equation}
\hat P_{s}(i,j)=\begin{cases} 
P_s(i,j) & {j \in top_k(i)} \\ 0& otherwise 
\label{local_transition}
\end{cases}
\end{equation}
where the ``local'' refers to the $k$-nearest neighbors. With $\hat P_{s}$, the transition probability of two distant points is set to zero.

\begin{figure}[!t]
\centering
        \includegraphics[width=1.6in]{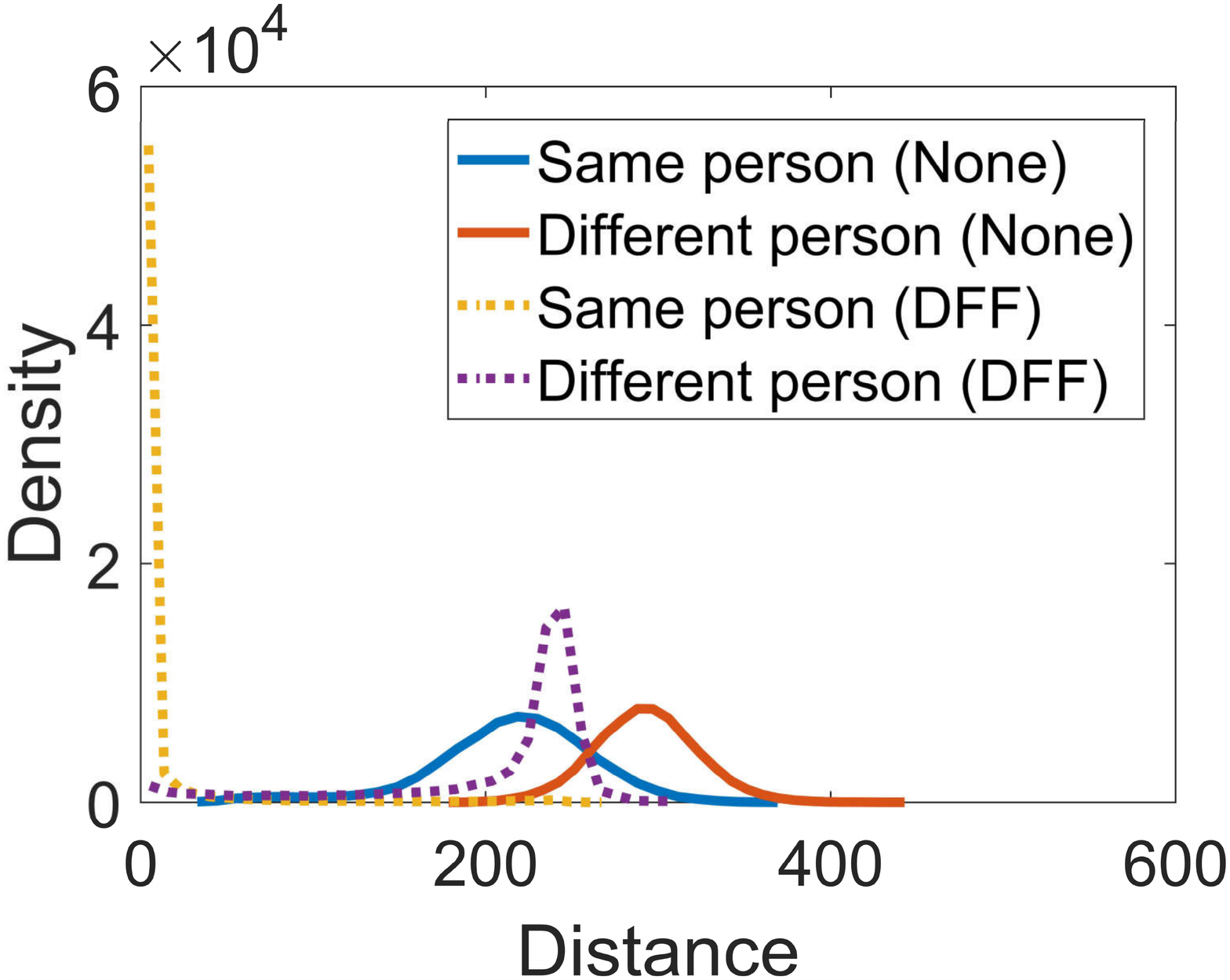} 
         
    \caption{
    The distance distributions before and after DFF on the Market-1501 dataset \cite{zheng2015scalable}. Baseline features: ID-discriminative Embedding \cite{Zheng2016Person} fine tuned on ResNet \cite{he2016deep}.  }
    \label{whyDFF} 
\end{figure}

In the second stage, multiple graphs are fused with a mixture Markov chain \cite{wang2012unsupervised}. The fusion process is essentially a random walk on multiple graphs in an iterative manner defined as:
\begin{equation}
P_s^{t+1}=\hat P_s \times (\frac{1}{S-1} \sum_{s' \neq s} P_{s'}^t) \times (\hat P_s)^T
\label{graphF}
\end{equation}
where $\hat P_s$ is viewed as a transition matrix, and the part in parentheses is viewed as an average status matrix after $t$ times' iteration. $P_s^{t+1}$ is the status matrix under $s$-th sub-feature after $t+1$ times' iteration. The above equation can be understood as an information exchange process among multiple graphs. After $t$ iterations, the overall status matrix is calculated as:
\begin{equation}
P^*=\frac{1}{S}\sum_{s=1}^{S}P_s^t
\end{equation}
The final distance metric used for re-ID evaluations is computed as $D=\frac {1}{P^*}$.

Fig.~\ref{whyDFF} shows the distance distribution of same person vs. different person, with and without DFF. Ideally, the distance between the same person should be as small as possible, preferably close to zero, while the distance between different persons should be larger. As can be seen in Fig.~\ref{whyDFF}, the distance distributions with DFF is significantly improved.

\begin{figure}[!t]
\centering
\subfigure[]{
        \includegraphics[width=1.55in]{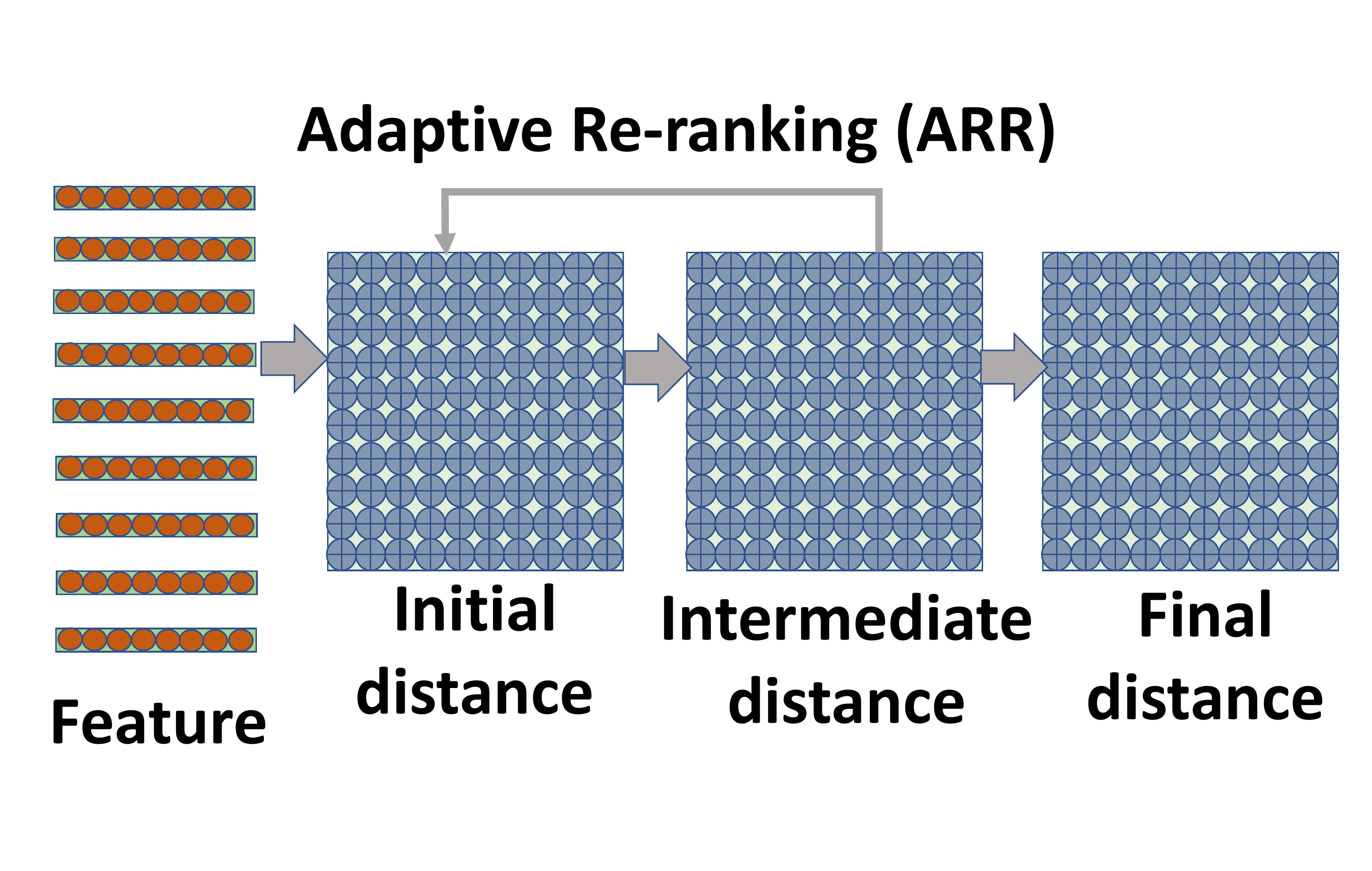} 
        \label{evaARR}  
          }
\subfigure[]{
        \includegraphics[width=1.55in]{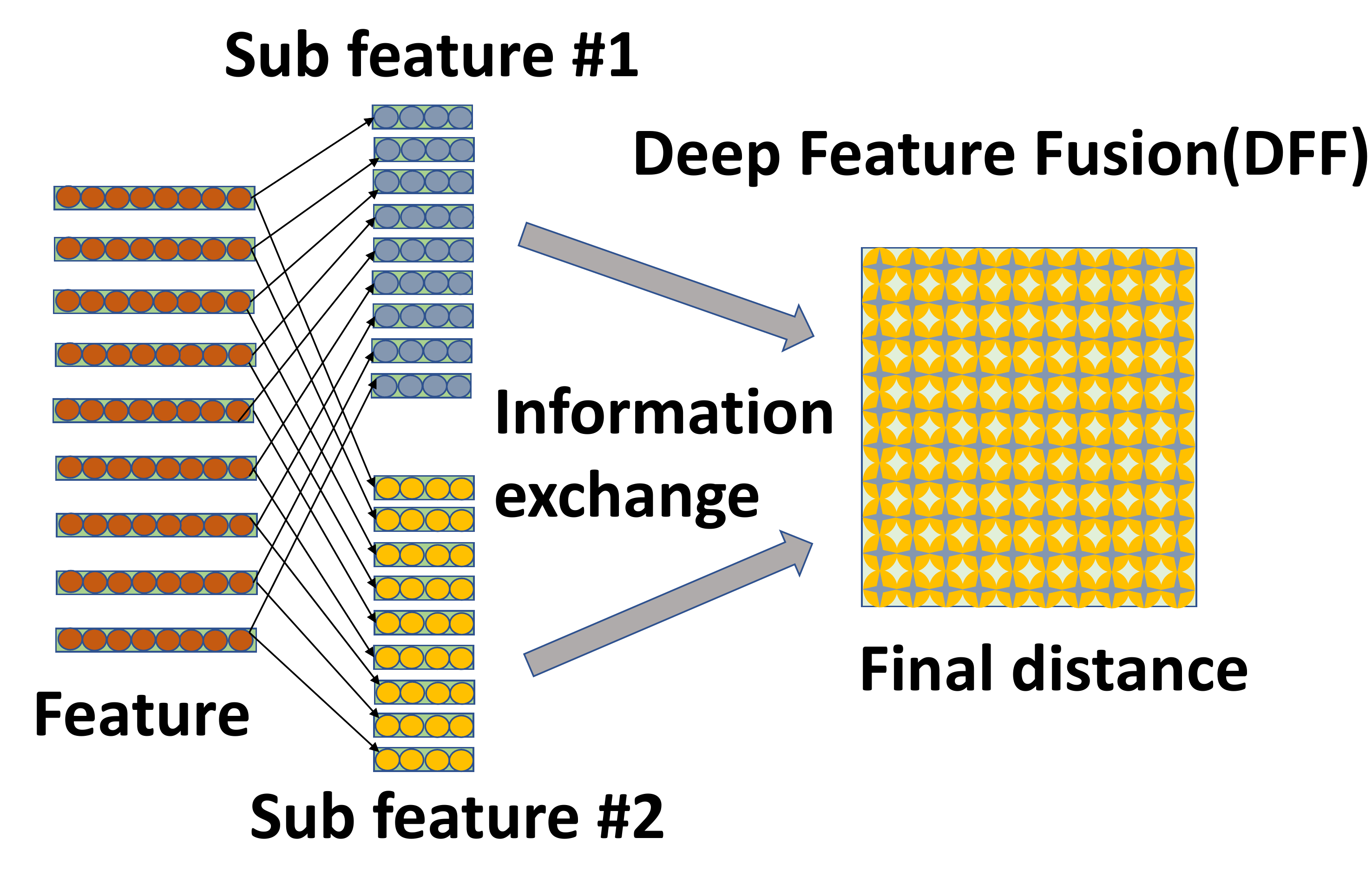}   
        \label{evaDFF}  
      }
    \caption{\label{evaARRDFF}(a) Adaptive Re-ranking (ARR). (b) Deep Feature Fusion (DFF). }
\end{figure}

\subsection{Combination of DFF and ARR}
\begin{figure}[htbp]
\centering
\subfigure[DFF-ARR.]{
        \includegraphics[width=1.55in]{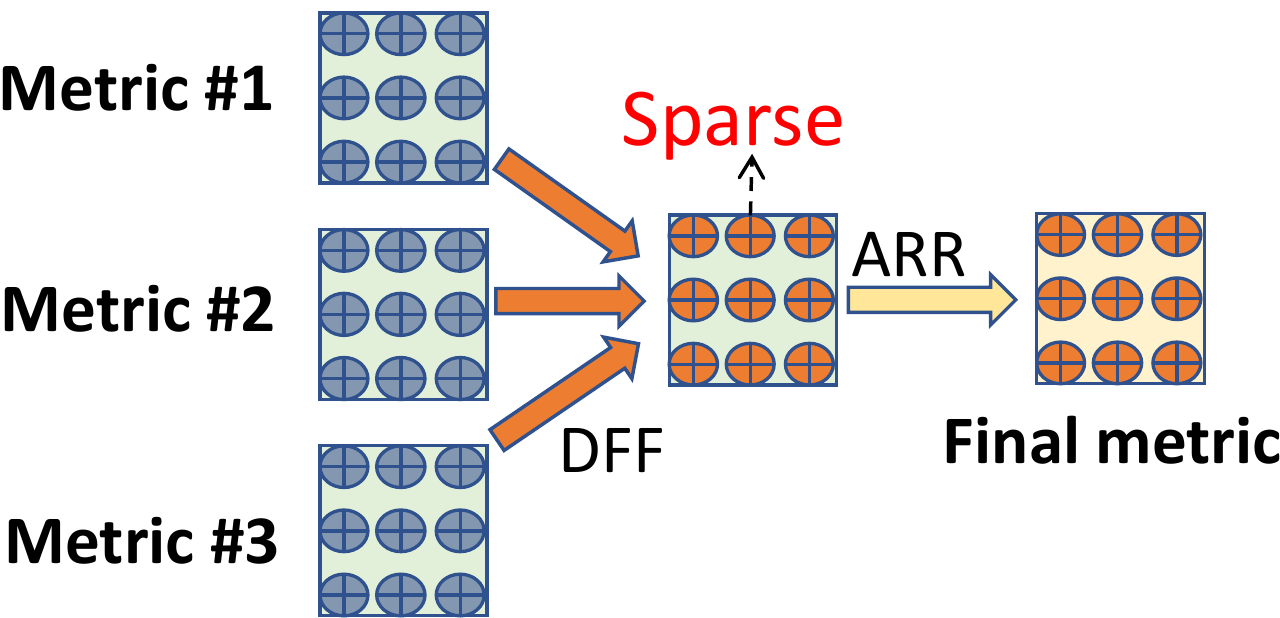}   
         \label{DFF-ARR}
      }
\subfigure[ARR-DFF.]{
        \includegraphics[width=1.55in]{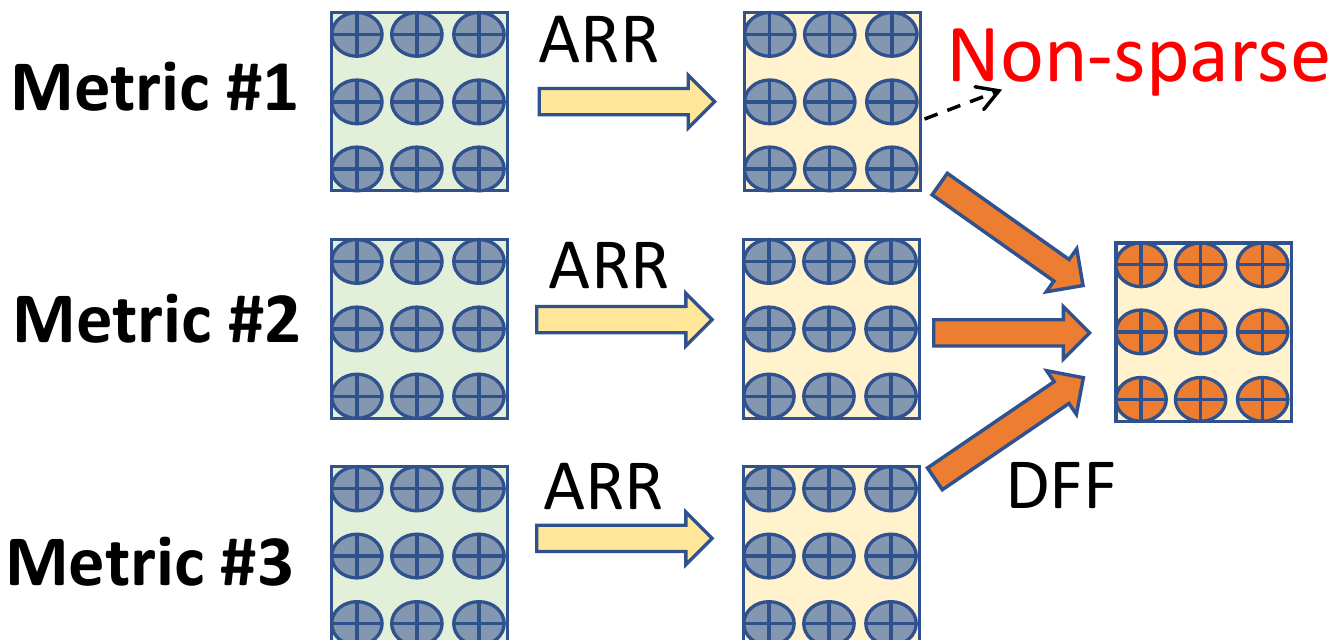} 
       \label{ARR-DFF}
          }
\caption{Combination strategies of ARR and DFF when $S$=3. (a) Metrics are first fused and then re-ranked. (b) Metrics are first re-ranked and then fused.}
\label{ARRDFF}
\end{figure}

Both DFF and ARR can be directly deployed into re-ID systems as shown in Fig.~\ref{evaARRDFF} or combined together. Fig.~\ref{ARRDFF} illustrates two strategies for combinations assuming each image's feature is evenly split into $3$ sub-features (e.g. $S=3$):

\begin{itemize}
\item DFF-ARR. Three metrics are first fused by the DFF into a single one, which is then re-defined by the ARR.
\item ARR-DFF. Each of the three distance metrics is first re-defined by the ARR. Then three re-defined metrics are fused into a single one by the DFF for the final evaluation.
\end{itemize}

Both combination strategies are evaluated although we expect the second strategy, ARR-DFF, would perform better. DFF-ARR has poor performance since the output matrix of DFF is ``sparse'' (Fig.~\ref{DFF-ARR}). In DFF-ARR, multiple metrics are first fused into a single one by DFF, denoted as $O$ which is a ``sparse'' matrix. The ``sparse'' here does not mean that most of the elements in $O$ are zero. Instead, they are quite large, almost equivalent to an infinite value. Therefore, $O$ can be viewed as a sparse matrix in a sense. The reason for a sparse $O$ lies in the fact that DFF employs the diffusion process which spreads the affinity information ``locally'' (see Eq.~\ref{local_transition}) and sets the distance of two distant points to infinity. Therefore, the distance information which is out of the close neighbors is eliminated in $O$. Directly inputting $O$ to the ARR may impair the performance because a lot of information has bee discarded. While in ARR-DFF (Fig.~\ref{ARR-DFF}), the output of ARR is ``non-sparse''. ARR utilizes the original Euclidean distance $D_E$ to compensate for the lack of information in $D_J$ which is also ``sparse'' (see Eq.~\ref{finaldist}).

\section{Experiments}

\subsection{Datasets and Settings}

{\bfseries Datasets.}Three commonly used benchmarks, namely Market-1501 (Market) \cite{zheng2015scalable}, DukeMTMC-reID (Duke) \cite{ristani2016performance}  and CUHK03 \cite{li2014deepreid}, are used in our evaluations. We report single-query evaluations for them.

The Market-1501 (Market) dataset contains 32,668 images of 1,501 identities detected by Deformable Part Model (DPM) \cite{felzenszwalb2008discriminatively} under six cameras. The dataset is divided into a training set with 12,936 images of 751 identities and a testing set with 19,732 images of 750 identities. In testing, 3,368 images of 750 identities are used as the probe set. 
The DukeMTMC-reID (Duke) dataset consists of 1,404 identities captured by 8 cameras. The training set contains 16,522 images of 702 identities, the testing set 17,661 images of 702 identities, and the probe set 2,228 images. It is one of the most challenging re-ID datasets up to now. 
The CUHK03 dataset contains 13,164 images of 1,360 identities under two cameras. CUHK03 provides both manually labeled and DPM-detected bounding boxes. We use the latter in this paper as it is more challenging. We follow the new evaluation protocol \cite{Zhong2017Re} instead of the single-shot protocol \cite{li2014deepreid} since the new protocol is more consistent with practice. 

{\bfseries Evaluation metrics.} Two evaluation metrics are used. 
The first one is the standard Cumulated Matching Characteristics (CMC) curve \cite{Wang2007Shape}. The CMC curve shows the probability that a probe appears in different-sized candidate lists. We report the cumulated matching accuracy at rank-1, abbreviated as R-1. The second one is the mean average precision (mAP) \cite{zheng2015scalable}. It considers both the precision and the recall when multiple ground truths exist.

{\bfseries Feature representations.} Five different feature extractors are used in our evaluation. The ID-discriminative Embedding features \cite{Zheng2016Person} trained on CaffeNet \cite{krizhevsky2012imagenet} (abbreviated as IDE(C)) and ResNet-50 \cite{he2016deep} (abbreviated as IDE(R)) are used. In addition, we also evaluate features trained on the  AlignedReID (Aligned) \cite{zhang2017alignedreid}, the Pose-Sensitive Embedding (PSE) \cite{sarfraz2017pose} and the Beyond Part Model (PCB)\cite{sun2017beyond}. They generate feature vectors of different dimensions and manifest merit in re-ID benchmarks. Extracted features are available \footnote{https://drive.google.com/open?id=1HzsC2qfjKgxYW8SfQX1wbxkk\\7PFtcy56}.

\subsection{Evaluation of Adaptive Re-ranking (ARR)}
\begin{figure*}[t]
      \begin{minipage}[t]{0.245\linewidth} 
        \centering   
        \includegraphics[width=\linewidth,height=0.94\linewidth]{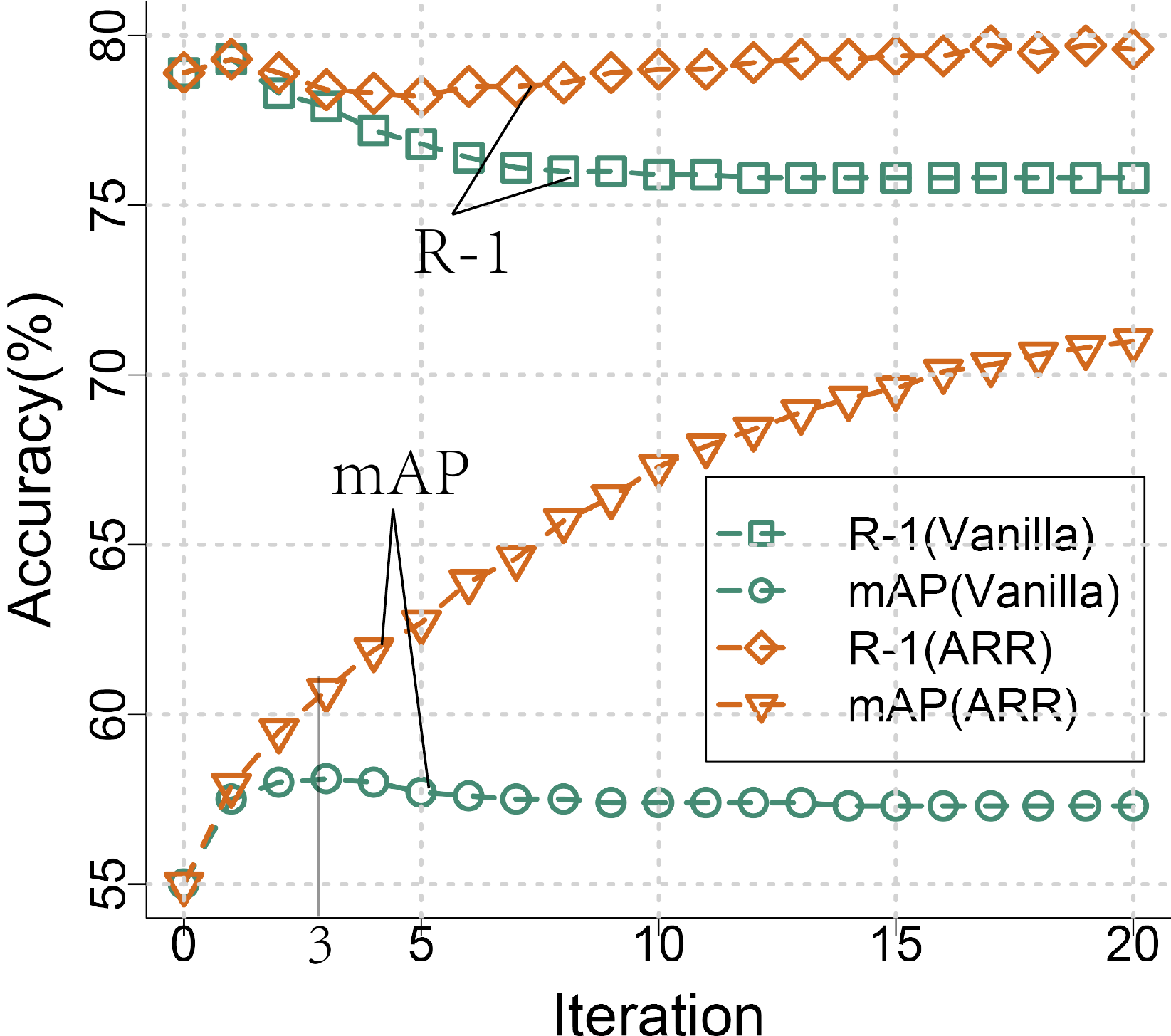}     
        \centerline{\small(a) $k=5$}
      \end{minipage}
      \begin{minipage}[t]{0.245\linewidth}   
        \centering   
        \includegraphics[width=\linewidth,height=0.94\linewidth]{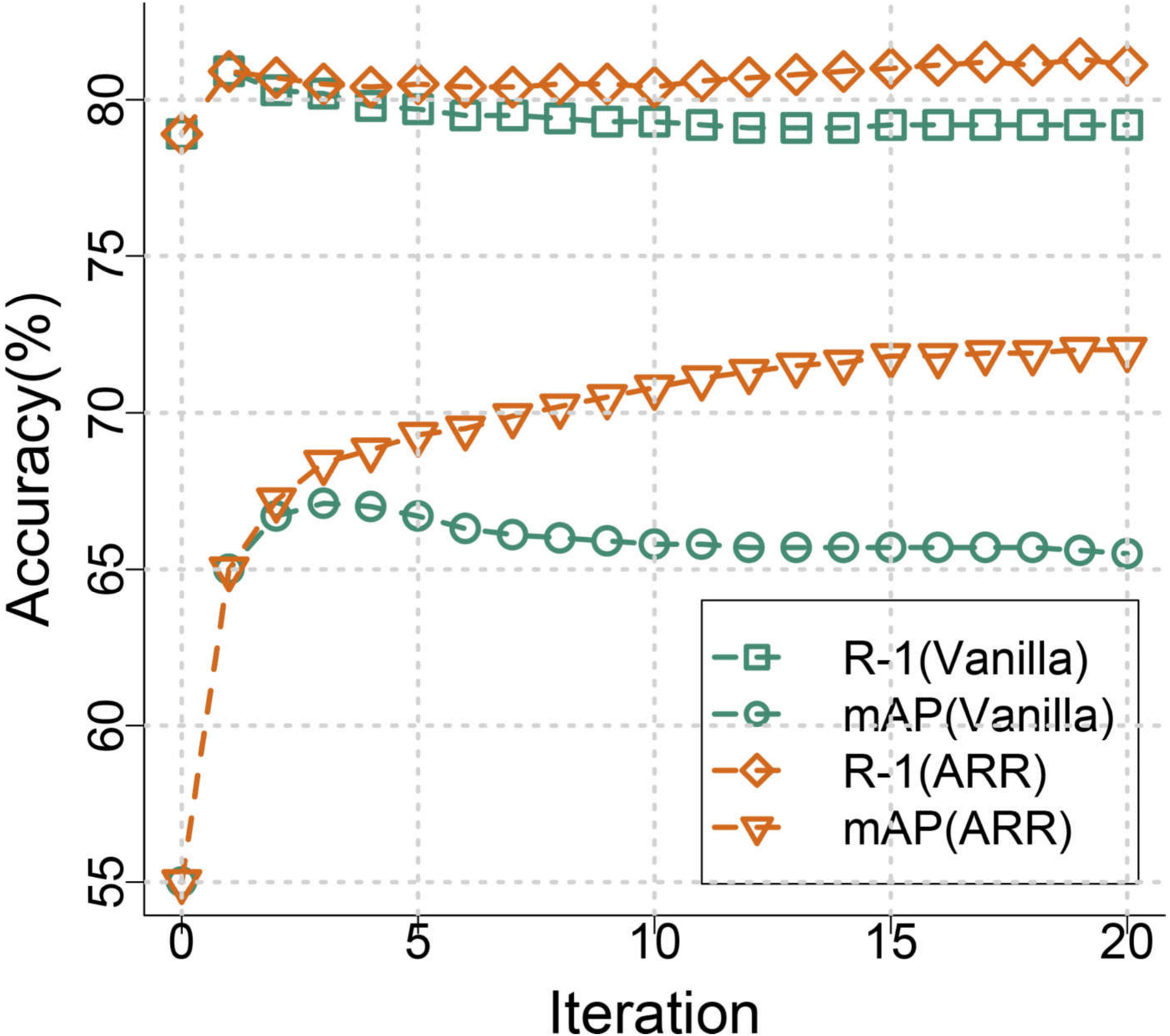}   
        \centerline{\small(b) $k=10$}
       \end{minipage} 
       \begin{minipage}[t]{0.245\linewidth}   
        \centering   
        \includegraphics[width=\linewidth,height=0.94\linewidth]{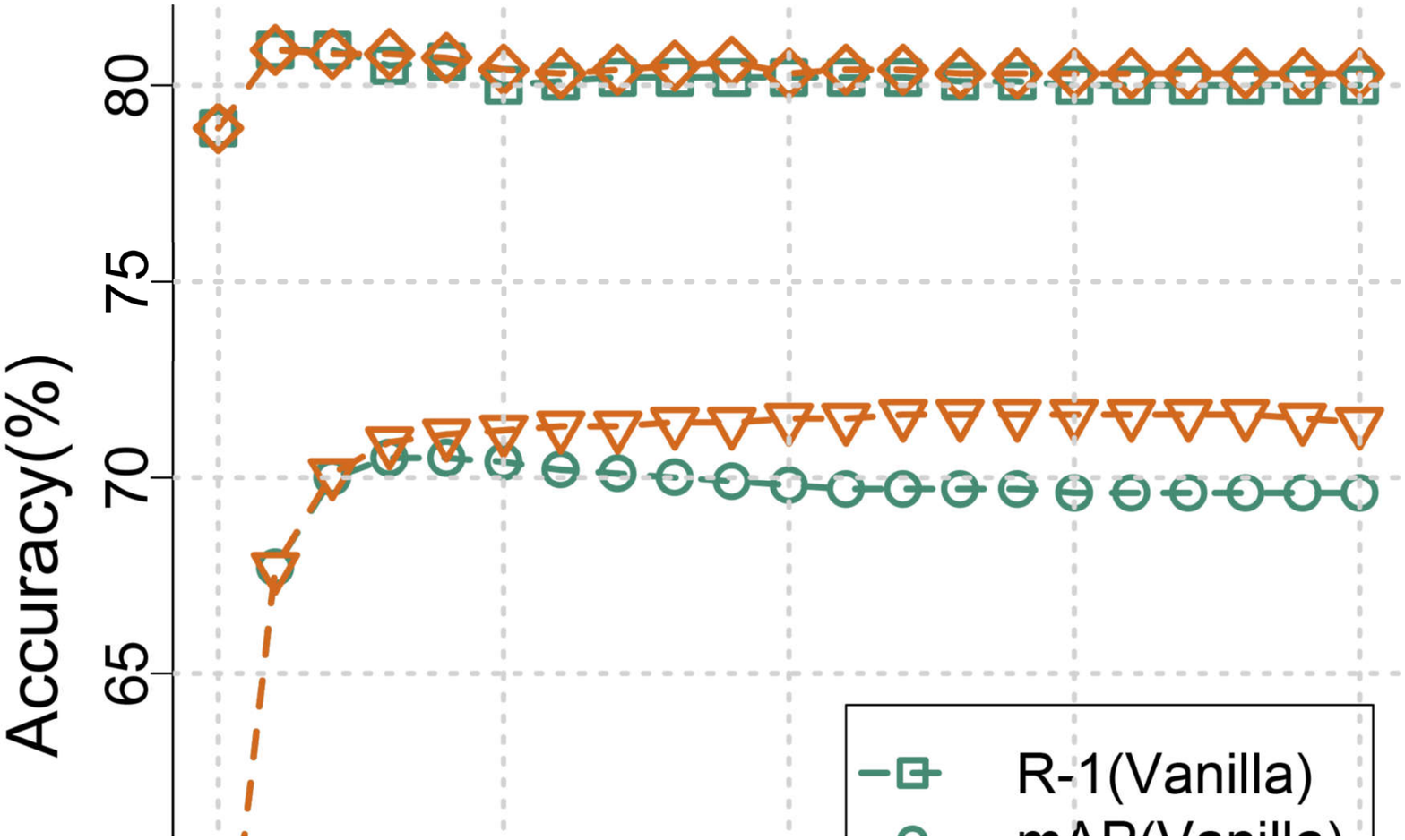}   
        \centerline{\small(c) $k=15$}
       \end{minipage} 
       \begin{minipage}[t]{0.245\linewidth}   
        \centering   
        \includegraphics[width=\linewidth,height=0.94\linewidth]{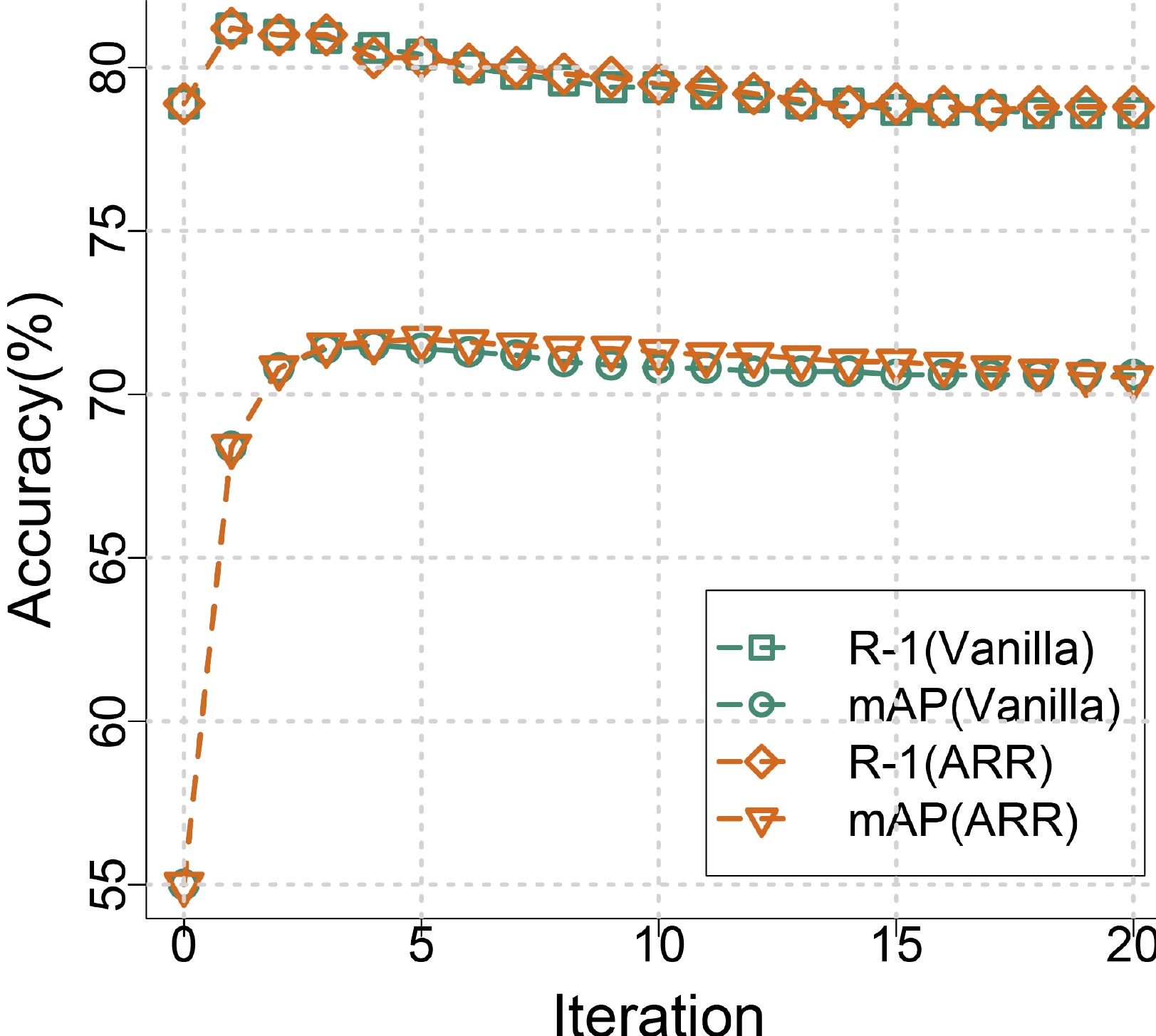}   
        \centerline{\small(d) $k=20$}
       \end{minipage}  
 \caption{The R-1 accuracy and mAP along iterations with different initial $k$: (a) $k=5$, (b) $k=10$, (c) $k=15$, (d) $k=20$ on Market. Baseline features: ID-discriminative Embedding (IDE) fine-tuned on ResNet.}
\label{iterReranking}
    \end{figure*}

Fig.~\ref{iterReranking} shows the analysis of effectiveness on Adaptive Re-ranking (ARR Fig.~\ref{evaARR}), on two datasets (Market and CUHK03) and two baselines (IDE(C) and IDE(R)). We first compare ARR with the vanilla version using both R-1 accuracy and mAP. The initial value $k$ is set as $k=5,10,15,20$ respectively. The baseline is the IDE features fine-tuned on ResNet. 

Fig.~\ref{iterReranking} shows that mAP of both two methods improve significantly through the iterative update process. But ARR is more robust than the vanilla version under different values of $k$. The performance of ARR is consistently better while the vanilla version degrades especially in the mAP under small $k$ ($k=5,10$). When $k$=5, the mAP of the vanilla version first rises along iterations, and then drops gradually along with the iteration. It reaches its peak point after the third iteration. In the case of ARR, mAP grows along iterations and outperforms the vanilla version by a considerable margin at low $k$. Actually the R-1 accuracy indicates the ability to retrieve the most similar match, while the mAP indicates the ability to find all true matches \cite{sun2017beyond}. Therefore, ARR is especially helpful in finding more matches. The performance of the two methods tends to be similar when $k$ becomes large. Note ARR does not require careful tuning for $k$.

We also compare the performance of ARR with other re-ranking methods on Market and CUHK03 datasets. The baselines are IDE(C) and IDE(R). We conduct a ten-iteration ARR and set $k$ to 15, 10 for Market and CUHK03 datasets respectively. Five widely used re-ranking algorithms for image retrieval and re-ID are compared, including contextual dissimilarity measure (CDM) \cite{Jegou2011A}, sparse contextual activation (SCA) \cite{bai2016sparse}, $k$-reciprocal encoding ($k$-RE) \cite{Zhong2017Re}, divide and fuse re-ranking (DaF) \cite{Yu2017Divide} and expanded cross neighborhood  (ECN) \cite{sarfraz2017pose}. The results are shown in Table~\ref{ARR}. As can be seen, ARR achieves competitive results on both two datasets in the R-1 accuracy and mAP. 

Compared to the baselines, ARR significantly improves the rank-1 accuracy and mAP. Especially in the mAP of Market dataset, our method gains an impressive increase of 16\% (48.9\%-32.9\%) on IDE(C) and 17.9\% (72.9\%-55.0\%) at the IDE(R) feature. That is because ARR utilizes the neighborhood structures to exploit the contextual information. Furthermore different from other single feature based re-ranking methods (CDM, SCA, $k$-RE and ECN) and multiple sub-features based method (DaF), ARR involves more contextual information along iterations automatically. Hence compared to the state-of-the-art method, ECN, ARR can achieve highly competitive performance.

\begin{table}[!t]
\small
 \centering
   \begin{tabular}{|c|c|c|c|c|c|c|c|c|}
    \hline
    \multirow{3}{*}{Method}&
    \multicolumn{2}{c|}{Market}&\multicolumn{2}{c|}{CUHK03}\cr\cline{2-5}
    &IDE(C)&IDE(R)&IDE(C)&IDE(R)\cr\cline{2-5}
    &R-1/mAP&R-1/mAP&R-1/mAP&R-1/mAP \cr\hline
    None&59.5/32.9&78.9/55.0&15.1/14.2&21.3/19.7\cr\hline
    CDM\cite{Jegou2011A}&60.1/42.5&79.8/56.7&16.8/17.5&22.9/20.6\cr\hline
    SCA\cite{bai2016sparse}&63.4/45.7&79.8/68.9&18.4/20.5&24.7/26.6\cr\hline
    k-RE\cite{Zhong2017Re}&63.4/47.2&81.4/70.4&21.3/19.7&24.9/27.3\cr\hline
    DaF\cite{Yu2017Divide}&64.3/48.2&{\bf 82.3}/72.4&19.3/21.4&26.4/30.0\cr\hline
    ECN\cite{sarfraz2017pose}&64.4/48.2&{\bf 82.3}/71.1&19.4/21.5&{\bf 27.3}/{\bf 30.2}\cr\hline
    {\bf ARR(ours)}&{\bf 64.8}/{\bf 48.9}&81.8/{\bf 72.9}&{\bf 21.7}/{\bf 21.9}&26.8/29.6\cr\hline
    \end{tabular}
    \caption{\label{ARR}Estimate ARR with other re-ranking methods}

\end{table}

\subsection{Evaluation of Deep Feature Fusion (DFF)}
\begin{table*}[!t]
 \centering
   \begin{tabular}{|c|c|c|c|c|c|c|c|c|}
    \hline
    \multirow{2}{*}{Dataset}&\multirow{2}{*}{Feature}&\multirow{2}{*}{Dimension}&
    \multicolumn{2}{c|}{None}&\multicolumn{2}{c|}{\bf DFF(ours)} &\multicolumn{2}{c|}{Gain of DFF}\cr\cline{4-9}
    &&&R-1&mAP&R-1&mAP&R-1&mAP \cr\hline
    \multirow{5}{*}{Market}&IDE(C) \cite{krizhevsky2012imagenet,Zheng2016Person}&4096&59.5&32.9&63.3&44.9&+3.8&
    {\color{red} +12.0}\cr\cline{4-9}
    &IDE(R) \cite{he2016deep,Zheng2016Person}&2048&78.9&55.0&80.6&68.0&+1.7&{\color{red} +13.0}\cr\cline{4-9}
    &PSE \cite{sarfraz2017pose}&1536&86.3&65.0&89.2&81.6&+2.9&{\color{red} +16.6}\cr\cline{4-9}
    &Alignedreid \cite{zhang2017alignedreid}&2048&89.2&75.9&90.4&85.3&+1.2&+9.4\cr\cline{4-9}
    &PCB \cite{sun2017beyond}&1536&92.9&78.5&91.8&82.2&-1.1&+3.7\cr\hline 
    \multirow{4}{*}{Duke}&IDE(R) \cite{he2016deep,Zheng2016Person}&2048&65.2&45.0&69.7&58.8&{\color{cyan}+4.5}&{\color{blue}+13.8}\cr\cline{4-9}
    &PSE \cite{sarfraz2017pose}&1536&79.8&62.0&84.2&76.9&{\color{cyan}+4.4}&{\color{blue}+14.9}\cr\cline{4-9}
    &Alignedreid \cite{zhang2017alignedreid}&2048&79.3&64.8&84.9&79.1&{\color{cyan}+5.6}&{\color{blue} +14.3}\cr\cline{4-9}
    &PCB \cite{sun2017beyond}&1536&84.5&70.0&86.3&75.8&+1.8&+5.8\cr\hline
    \multirow{4}{*}{CUHK03}&IDE(C) \cite{krizhevsky2012imagenet,Zheng2016Person}&1024&15.1&14.2&17.8&18.1&{\color{magenta}+2.7}&{\color{magenta}+3.9}\cr\cline{4-9}
    &IDE(R) \cite{he2016deep,Zheng2016Person}&2048&21.3&19.7&23.6&24.4&{\color{magenta}+2.3}&{\color{magenta}+4.7}\cr\cline{4-9}
    &PCB \cite{sun2017beyond}&1536&59.1&53.9&59.1&57.2&0.0&+3.3\cr\cline{4-9}
    &Alignedreid \cite{zhang2017alignedreid}&2048&60.7&56.0&68.7&68.8&{\color{magenta}+8.0}&{\color{magenta}+12.8}\cr\hline
    \end{tabular}
    \caption{\label{DFF}Performance of DFF at different features on three datasets.}
\end{table*}

The effectiveness of Deep Feature Fusion (DFF, Fig.~\ref{evaDFF}) is evaluated on Market, Duke and CUHK03 datasets.  The results are shown in Table ~\ref{DFF}. Experiments of each dataset are performed on at least four deep features including average ones (IDE(C) and IDE(R)) and superior ones (AlignedReID, PSE and PCB). Note that all the five features are high dimensional ranging from 1,024 dimensions to 4,096 dimensions. The baselines are the results from Euclidean distance without additional techniques. In the diffusion process of DFF, we conduct a one-iteration fusion with the number of sub-features $S=4$ throughout the experiments. 

As can be seen in Table ~\ref{DFF}, DFF significantly improve the performance on all datasets compared to the baselines. On the Market dataset, DFF has massive gains of 16.6\%, 13.0\% and 12.0\% improvement (marked red in Table ~\ref{DFF}) in the mAP at PSE, IDE(R) and IDE(C) features respectively. Actually, DFF improves the mAP in all cases on Market. In the R-1 accuracy, DFF consistently leads to improvement except the case of PCB feature. This particular performance degradation is caused by the characteristic of PCB features. In PCB, part-level features are concatenated to form the final descriptor during testing. Therefore, our DFF method may implicitly destroy the structural information of PCB features. 
The improvement in terms of mAP is larger than that in the R-1 accuracy for all five baselines. Recall that the mAP characterizes the ability to find all true matches. Different sub-features manifest different characteristics which can bring more ground truths and result in a higher mAP. On the Duke dataset, DFF leads to a large absolute improvement of 13.8-14.9\% (marked blue) in the mAP at IDE(R), PSE and Alignedreid features. In addition, DFF results in an improvement of 4.4-5.6\% (marked cyran) in the R-1 accuracy at the first three features. At the PCB feature, DFF still provides meaningful performance gains. In general, DFF yields consistent improvements at all baselines. On the CUHK03 dataset, we report four baselines. At IDE(C), IDE(R) and Alignedreid, the relative increase achieved by DFF ranges from 2.3\% to 8.0\% in the R-1 accuracy and 3.9\% to 12.8\% in the mAP (magenta marked). At the PCB feature, DFF leads no improvement in the R-1 accuracy but still increases the mAP by 3.3\%. 

In summary, experiments conducted with three datasets and five baselines verify the effectiveness of DFF. It is noticeable that DFF achieves large margins compared with the baselines especially in terms of mAP.  

\subsection{Evaluation of Adaptive Re-ranking of Deep Features}
Here we evaluates the performances of deep feature fusion with adaptive re-ranking combined, 
namely DFF-ARR and ARR-DFF, introduced in Section 3.3.

Table~\ref{ALL} lists the comparison results of ARR-DFF and DFF-ARR. For the parameters in our method, we set $k$ to 15, 10 for the Market and the CUHK03 dataset, respectively. Three-iteration ARR and one-iteration DFF with the sub-feature $S=4$ are used on three datasets.

\begin{figure}[htb]
\centering
\includegraphics[width=7.4cm,height=4cm]{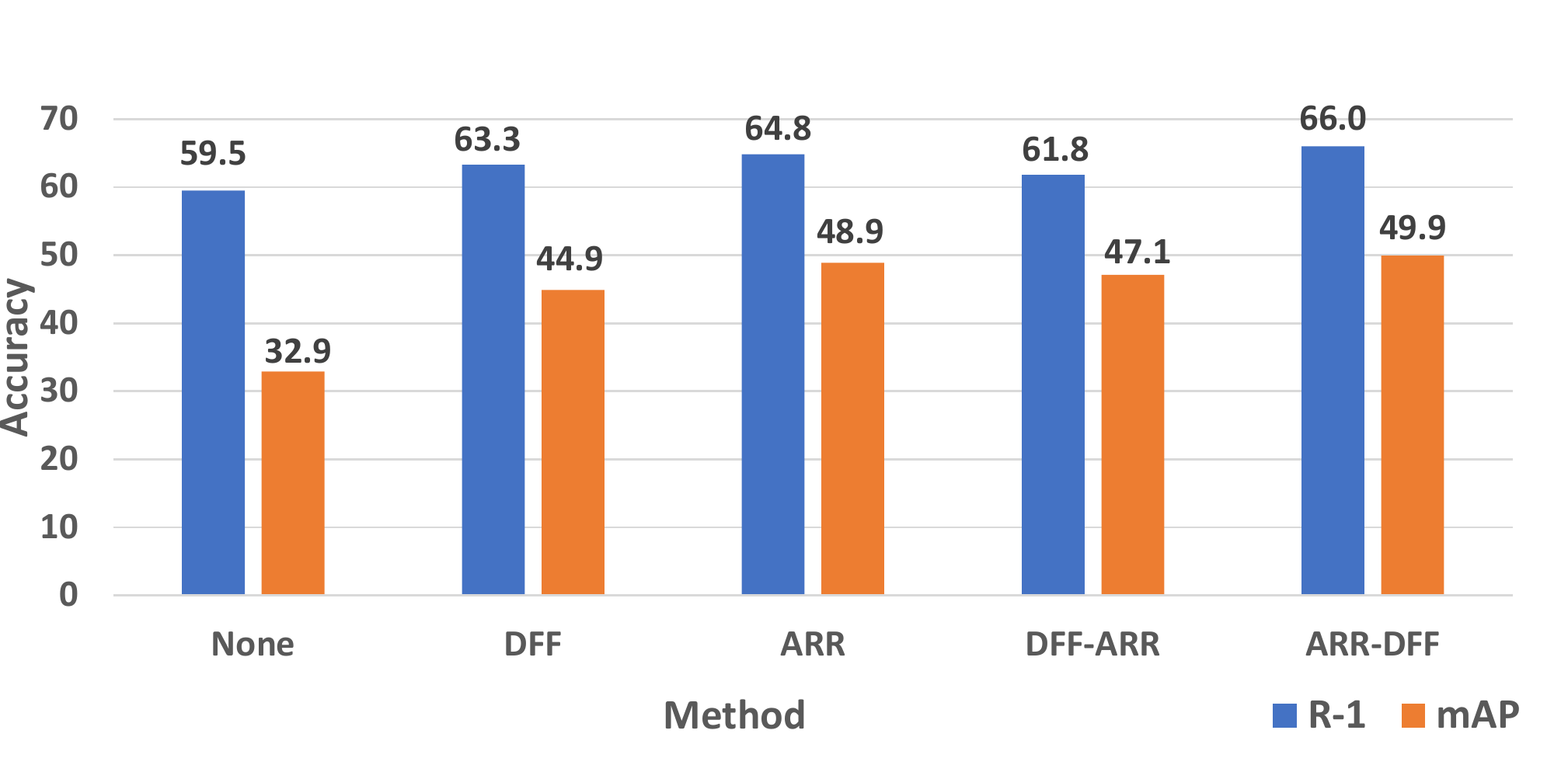} 
\caption{Performance of different methods including DFF, ARR, DFF-ARR and ARR-DFF at IDE(C) on Market dataset.}
\label{easydis}
\end{figure}

\begin{table}[htb]
\small
 \centering
   \begin{tabular}{|c|c|c|c|c|c|c|}
   \hline
    \multirow{2}{*}{Method}&\multicolumn{2}{c|}{Market}&\multicolumn{2}{c|}{Duke}&\multicolumn{2}{c|}{CUHK03}
    \cr \cline{2-7}
    &R-1&mAP&R-1&mAP&R-1&mAP \cr \hline
    IDE(C) \cite{krizhevsky2012imagenet, Zheng2016Person}&59.5&32.9&-&-&15.1&14.2 \cr 
    with DFF-ARR&61.8&47.1&-&-&18.4&20.5 \cr 
    {\bf with ARR-DFF}&{\bf 66.0}&{\bf 49.9}&-&-&{\bf 20.9}&{\bf 21.4} \cr \hline
    IDE(R) \cite{he2016deep,Zheng2016Person}&78.9&55.0&65.2&45.0&21.3&19.7 \cr
    with DFF-ARR&79.3&69.5&67.6&60.3&24.5&27.5 \cr
    {\bf with ARR-DFF}&{\bf 82.4}&{\bf 73.2}&{\bf 68.3}&{\bf 63.7}&{\bf 27.4}&{\bf 30.2} \cr \hline
    PSE \cite{sarfraz2017pose}&86.3&65.0&79.8&62.0&-&- \cr
    with DFF-ARR&87.9&82.3&81.5&78.0&-&- \cr
    {\bf with ARR-DFF}&{\bf 90.6}&{\bf 84.6}&{\bf 85.5}&{\bf 78.3}&-&- \cr \hline
    PCB \cite{sun2017beyond}&{\bf 92.9}&78.5&84.5&70.0&59.1&53.9 \cr
    with DFF-ARR&91.1&{\bf 86.4}&84.9&{\bf 81.3}&{\bf 65.1}&{\bf 67.3} \cr
    {\bf with ARR-DFF}&91.8&85.6&{\bf 86.4}&80.2&63.5&64.2 \cr \hline
    Alignedreid \cite{zhang2017alignedreid}&89.2&75.9&79.3&64.8&60.7&56.0 \cr
    with DFF-ARR&89.0&84.9&82.3&79.1&70.4&72.5 \cr
	{\bf with ARR-DFF}&{\bf 91.5}&{\bf 87.6}&{\bf 85.0}&{\bf 80.8}&{\bf 71.6}&{\bf 74.2} \cr \hline
    \end{tabular}
    \caption{\label{ALL}Performance of DFF-ARR and ARR-DFF at different features on three datasets. }
\end{table}
As shown in Table~\ref{ALL}, DFF-ARR and ARR-DFF has considerable gains compared with the baselines on all datasets. It can be seen that the gains from ARR-DFF are greater than that of DFF-ARR on R-1 accuracy and mAP. To further illustrate that, we demonstrate the performance of DFF, ARR, DFF-ARR and ARR-DFF at IDE(C) on Market dataset in one graph as shown in Fig.~\ref{easydis}. 

We can see that ARR-DFF generally outperforms DFF-ARR on the R-1 accuracy (66.0 \% vs 61.8\%) and mAP (49.9\% vs 47.1\%). In addition, DFF-ARR impairs the performance slightly as analyzed in Section 3.3. This observation generally holds in other cases. PCB is a different case as DFF may damage the structural information of PCB features as analyzed in Section 4.3.

In Table~\ref{SOA}, we also compare the performance of ARR-DFF with other state-of-the-art methods on three datasets. Our proposed method achieves best performance on most of them.
Specifically, our method exceeds SPreID \cite{Kalayeh2018Human} by 4.2\% (87.6\%-83.4\%) and 7.5\% (80.8\%-73.3\%) in the mAP on the Market and Duke datasets respectively. On the CUHK03	dataset, ARR-DFF obtains remarkable R-1 accuracy at 71.6\% and mAP at 74.2\%.

\begin{table}[!t]
\small
 \centering
   \begin{tabular}{|c|c|c|c|c|c|c|}
   \hline
    \multirow{2}{*}{Method}&\multicolumn{2}{c|}{Market}&\multicolumn{2}{c|}{Duke}&\multicolumn{2}{c|}{CUHK03}\cr \cline{2-7}
    &R-1&mAP&R-1&mAP&R-1&mAP \cr\hline
    ACRN \cite{Schumann2017Person}&83.6&62.6&72.6&52.0&-&- \cr\hline
    SVDNet \cite{Sun2017SVDNet}&82.3&62.1&76.7&56.8&-&-\cr\hline
    AACN \cite{Xu2018Attention}&85.9&66.9&76.8&59.3&-&-\cr\hline
    PAN \cite{zheng2017pedestrian}&86.7&69.3&71.6&51.5&36.3&34.0  \cr\hline
    HA-CNN \cite{Li2018Harmonious}&91.2&75.7&80.5&63.8&41.7&38.6 \cr\hline
    AWTL \cite{Ristani2018Features} &89.5&75.7&79.8&63.4&-&- \cr\hline
    MLFN \cite{Chang2018Multi}&90.0&74.3&81.2&62.8&52.8&47.8  \cr\hline
    DuATM \cite{Si2018Dual}&91.4&76.6&81.8&64.6&-&-\cr\hline
    Part-aligned \cite{Suh2018Part}&91.7&79.6&84.4&69.3&-&-\cr\hline
    SPreID \cite{Kalayeh2018Human}&{\bf 93.7}&83.4&{\bf 86.0}&73.3&-&-\cr\hline
    {\bf ARR-DFF}&91.5&{\bf 87.6}&85.0&{\bf 80.8}&{\bf 71.6}&{\bf 74.2}\cr\hline
    \end{tabular}
    \caption{\label{SOA}Comparison to state-of-the-art on three datasets.}
\end{table}

\subsection{Analysis of Parameters}

In sub-features splitting, the parameter $S$ represents the number of sub-features. In adaptive re-ranking, the parameter $k$ is the number of nearest neighbors which controls the amount of contextual information. Selecting $S$ and $k$ is an important issue when using ARR-DFF and we analyse the two parameters experimentally in this subsection. The baselines are IDE(R) (an average baseline) and Alignedreid (a superior baseline) features on the challenging CUHK03 dataset.

\begin{figure}[!htb]
\centering
\subfigure[R-1 under different $S$.]{
        \includegraphics[width=1.55in]{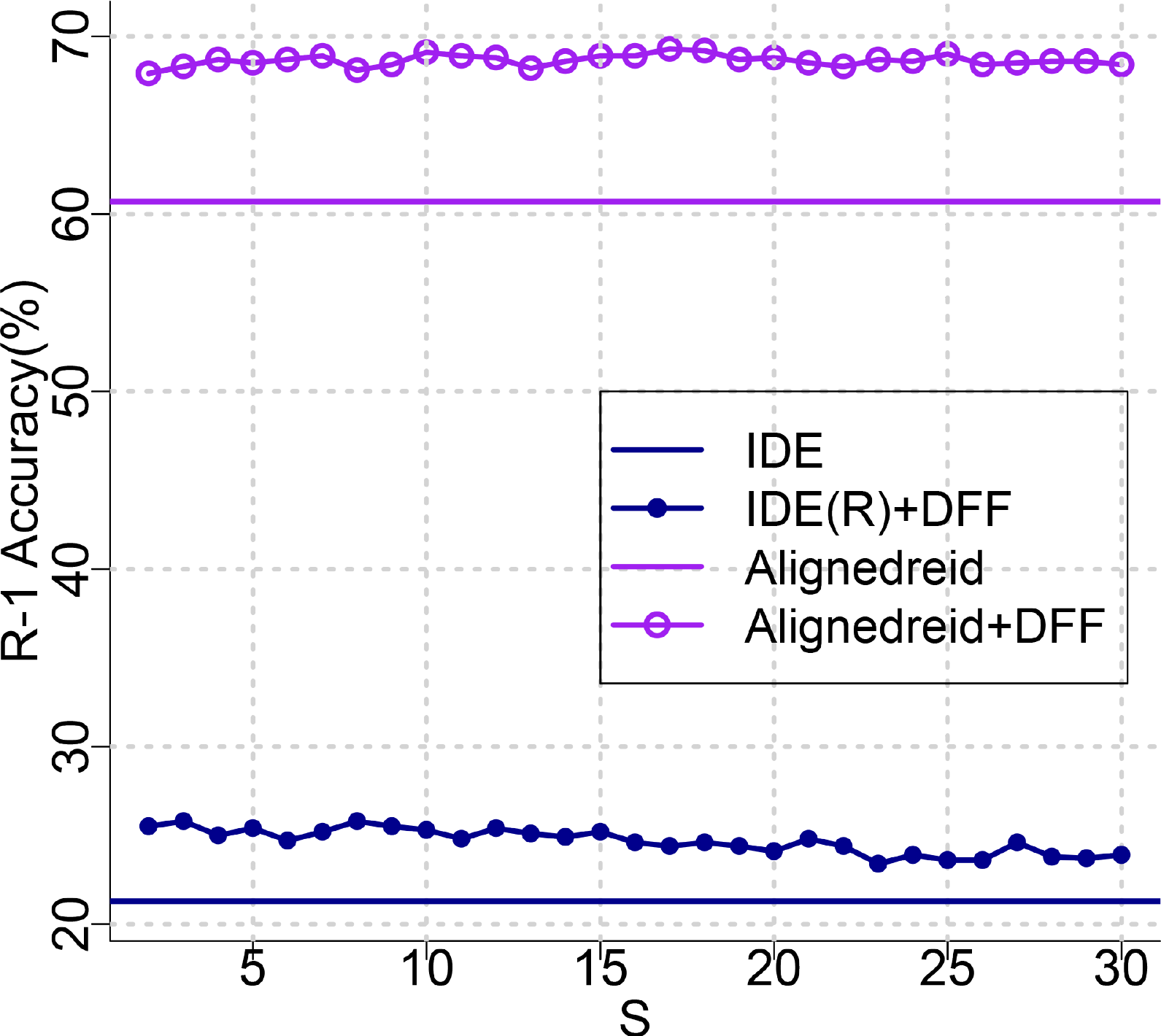}   
        \label{11}  
      }
\subfigure[mAP under different $S$.]{
        \includegraphics[width=1.55in]{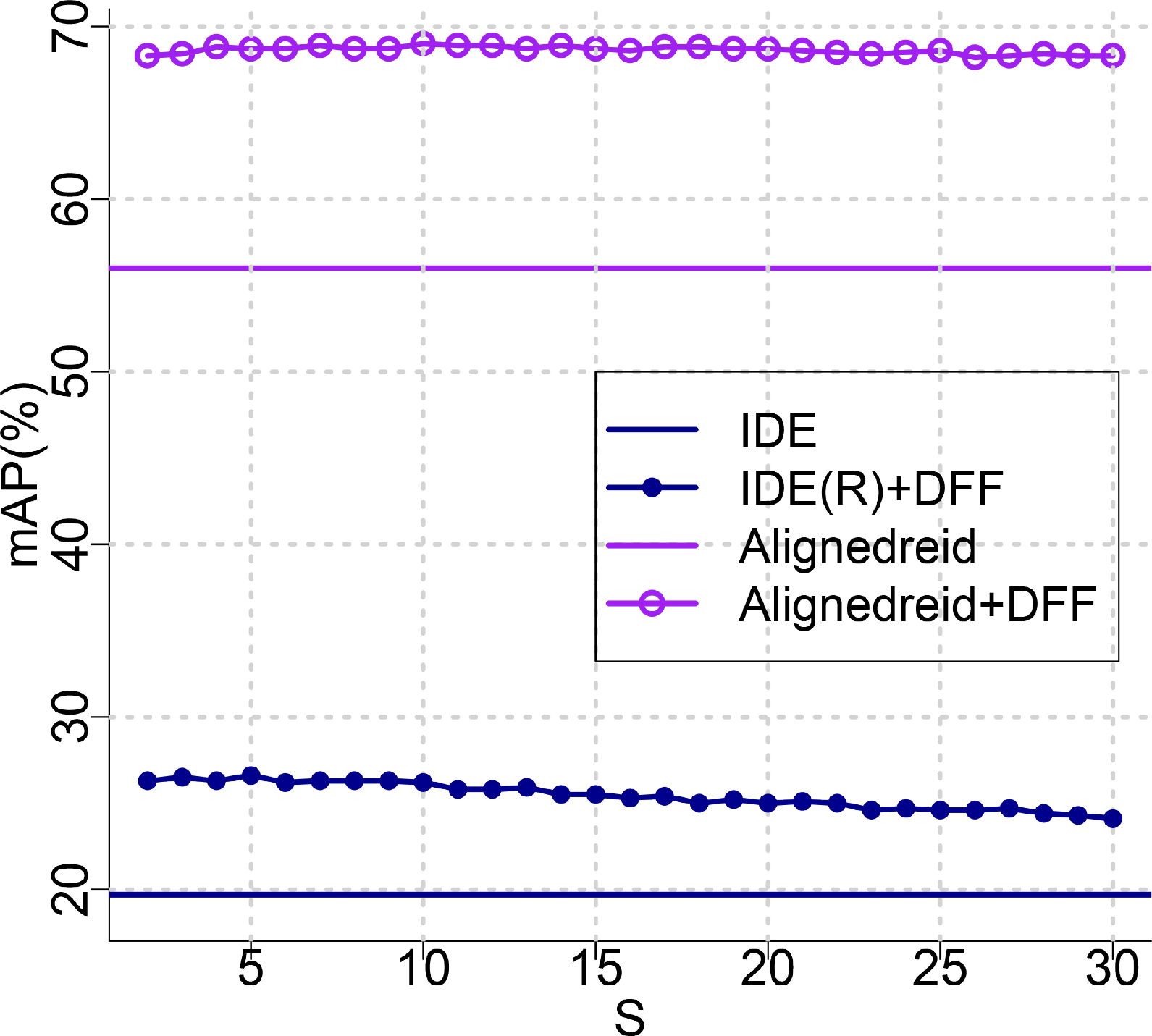} 
        \label{22}  
          }
\subfigure[R-1 under different $k$.]{
        \includegraphics[width=1.55in]{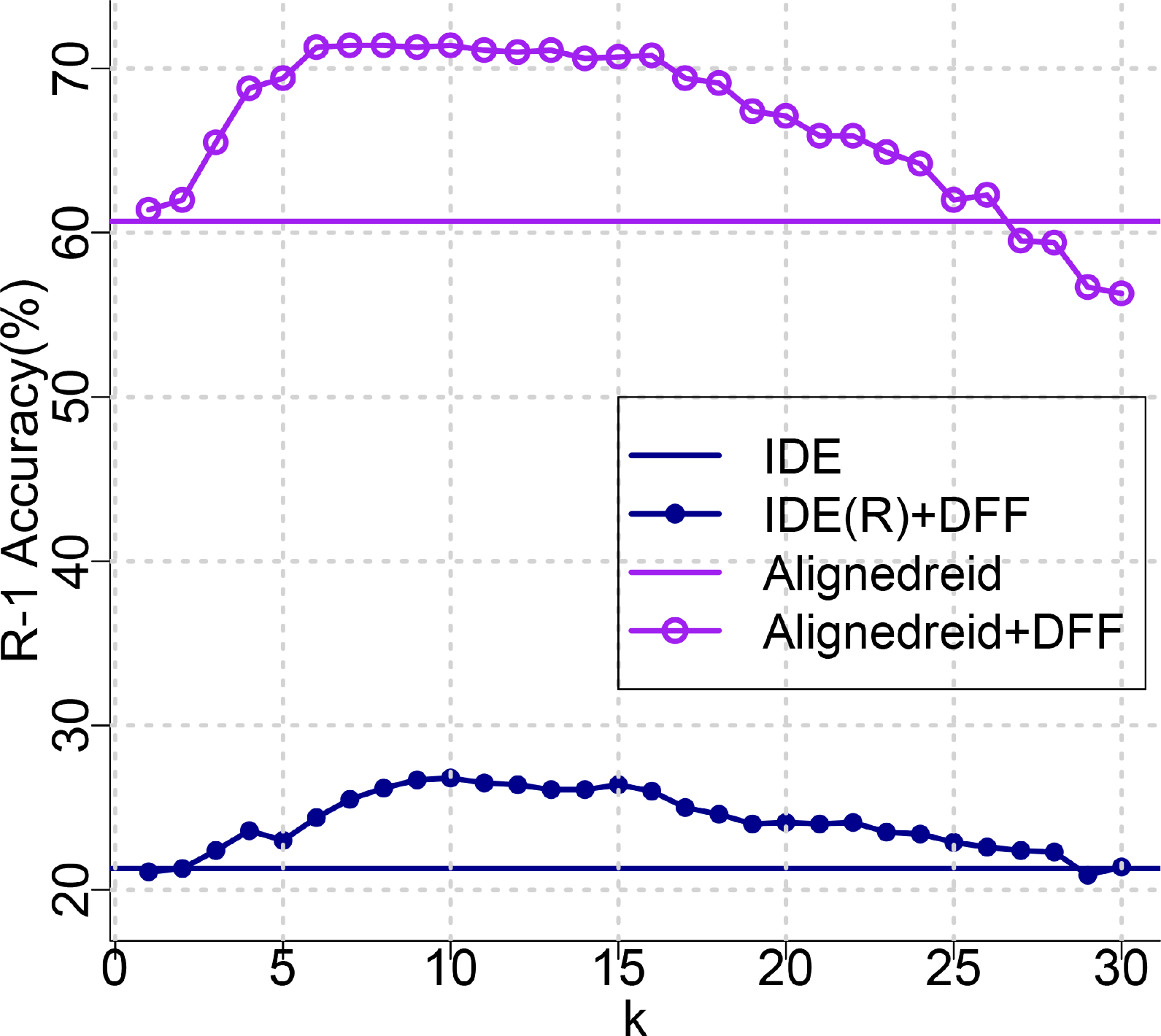}   
        \label{33}  
      }
\subfigure[mAP under different $k$.]{
        \includegraphics[width=1.55in]{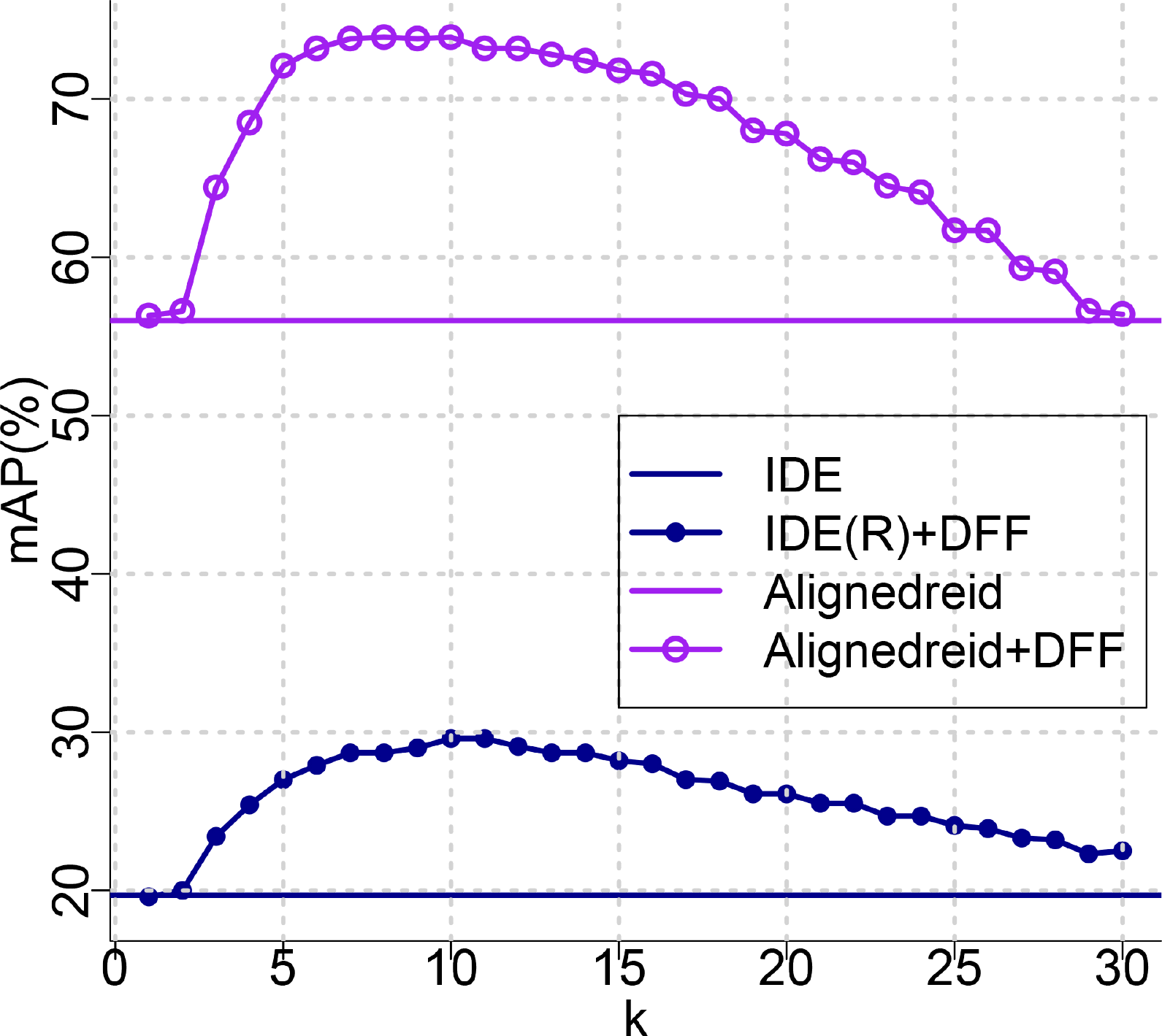} 
        \label{44}  
          }
    \caption{The impact of the parameters $S$ and $k$ on re-ID performance on CUHK03 dataset.}
\label{para}
\end{figure}

The impact of $S$ is shown in Fig.~\ref{11}, Fig.~\ref{22}. We can see that DFF consistently outperforms the baseline at different features. As $S$ increases, the R-1 accuracy and mAP at IDE(R) feature show a slowly declining trend but generally stable. The performance at Alignedreid feature is also robust under different $S$. 

The impact of $k$ is shown in Fig.~\ref{33}, Fig.~\ref{44}.  Similar trends can be observed on different features, rising at first and then returning gradually. When $k$ is set to 1, the re-ranking method exploits no contextual information and the result is similar to the baseline result. When $k$ is large, the contextual information which may become superfluous as noises are included so leading to a decline in performance.

\section{Conclusion}

This paper aims to improve re-ID, mainly by better utilizing deep learned features. 
First, an adaptive re-ranking (ARR) is proposed based on existing iterative re-ranking method to increase the use of contextual information during the iterations. In this way, non-relevant images are removed and more matches can be considered. Second, a deep feature fusion (DFF) method is proposed, which integrates metrics on  sub-features to improve overall improvements. Our study shows that both ARR and DFF are indeed effective in term of improving re-ID accuracies on different datasets using different features.  Moreover the combination of these two using the ARR-DFF strategy can bring further improvement.  In addition, our study shows that our methods are not sensitive to parameters.  Overall adaptive re-ranking with deep feature fusion can outperform state-of-art methods and serve as a competitive candidate to be plugged into existing re-ID systems.

{\small
\bibliographystyle{ieee}
\bibliography{egbib}

\begin{thebibliography}{10}\itemsep=-1pt

\bibitem{bai2016sparse}
S.~Bai and X.~Bai.
\newblock Sparse contextual activation for efficient visual re-ranking.
\newblock {\em IEEE Transactions on Image Processing}, 25(3):1056--1069, 2016.

\bibitem{Chang2018Multi}
X.~Chang, T.~M. Hospedales, and T.~Xiang.
\newblock Multi-level factorisation net for person re-identification.
\newblock In {\em CVPR}, 2018.

\bibitem{felzenszwalb2008discriminatively}
P.~Felzenszwalb, D.~McAllester, and D.~Ramanan.
\newblock A discriminatively trained, multiscale, deformable part model.
\newblock In {\em CVPR}, 2008.

\bibitem{Garcia2015Person}
J.~Garcia, N.~Martinel, C.~Micheloni, and A.~Gardel.
\newblock Person re-identification ranking optimisation by discriminant context
  information analysis.
\newblock In {\em ICCV}, 2015.

\bibitem{gong2014person}
S.~Gong, M.~Cristani, S.~Yan, and C.~C. Loy.
\newblock {\em Person re-identification}.
\newblock Springer, 2014.

\bibitem{he2016deep}
K.~He, X.~Zhang, S.~Ren, and J.~Sun.
\newblock Deep residual learning for image recognition.
\newblock In {\em CVPR}, 2016.

\bibitem{Hirzer2012Relaxed}
M.~Hirzer, P.~M. Roth, M.~Stinger, and H.~Bischof.
\newblock Relaxed pairwise learned metric for person re-identification.
\newblock In {\em ECCV}, 2012.

\bibitem{Jegou2011A}
H.~Jegou, H.~Harzallah, and C.~Schmid.
\newblock A contextual dissimilarity measure for accurate and efficient image
  search.
\newblock In {\em CVPR}, 2007.

\bibitem{Kalayeh2018Human}
M.~M. Kalayeh, E.~Basaran, M.~Gokmen, M.~E. Kamasak, and M.~Shah.
\newblock Human semantic parsing for person re-identification.
\newblock In {\em CVPR}, 2018.

\bibitem{krizhevsky2012imagenet}
A.~Krizhevsky, I.~Sutskever, and G.~E. Hinton.
\newblock Imagenet classification with deep convolutional neural networks.
\newblock In {\em NIPS}, 2012.

\bibitem{Leng2015Person}
Q.~Leng, R.~Hu, C.~Liang, Y.~Wang, and J.~Chen.
\newblock Person re-identification with content and context re-ranking.
\newblock {\em Multimedia Tools and Applications}, 74(17):6989--7014, 2015.

\bibitem{li2014deepreid}
W.~Li, R.~Zhao, T.~Xiao, and X.~Wang.
\newblock Deepreid: Deep filter pairing neural network for person
  re-identification.
\newblock In {\em CVPR}, 2014.

\bibitem{Li2018Harmonious}
W.~Li, X.~Zhu, and S.~Gong.
\newblock Harmonious attention network for person re-identification.
\newblock In {\em CVPR}, 2018.

\bibitem{Liao2015Person}
S.~Liao, Y.~Hu, X.~Zhu, and S.~Z. Li.
\newblock Person re-identification by local maximal occurrence representation
  and metric learning.
\newblock In {\em CVPR}, 2015.

\bibitem{luo2013shape}
L.~Luo, C.~Shen, C.~Zhang, and A.~van~den Hengel.
\newblock Shape similarity analysis by self-tuning locally constrained
  mixed-diffusion.
\newblock {\em IEEE Transactions on Multimedia}, 15(5):1174--1183, 2013.

\bibitem{Matsukawa2016Hierarchical}
T.~Matsukawa, T.~Okabe, E.~Suzuki, and Y.~Sato.
\newblock Hierarchical gaussian descriptor for person re-identification.
\newblock In {\em CVPR}, 2016.

\bibitem{mirmahboub2018person}
B.~Mirmahboub, M.~L. Mekhalfi, and V.~Murino.
\newblock Person re-identification by order-induced metric fusion.
\newblock {\em Neurocomputing}, 275:667--676, 2018.

\bibitem{pedronette2011exploiting}
D.~C.~G. Pedronette and R.~da~S~Torres.
\newblock Exploiting contextual spaces for image re-ranking and rank
  aggregation.
\newblock In {\em ICMR}, 2011.

\bibitem{qin2011hello}
D.~Qin, S.~Gammeter, L.~Bossard, T.~Quack, and L.~Van~Gool.
\newblock Hello neighbor: Accurate object retrieval with k-reciprocal nearest
  neighbors.
\newblock In {\em CVPR}, 2011.

\bibitem{rahman2011learning}
M.~M. Rahman, S.~K. Antani, and G.~R. Thoma.
\newblock A learning-based similarity fusion and filtering approach for
  biomedical image retrieval using svm classification and relevance feedback.
\newblock {\em IEEE Transactions on Information Technology in Biomedicine},
  15(4):640--646, 2011.

\bibitem{ristani2016performance}
E.~Ristani, F.~Solera, R.~Zou, R.~Cucchiara, and C.~Tomasi.
\newblock Performance measures and a data set for multi-target, multi-camera
  tracking.
\newblock In {\em ECCV}, 2016.

\bibitem{Ristani2018Features}
E.~Ristani and C.~Tomasi.
\newblock Features for multi-target multi-camera tracking and
  re-identification.
\newblock In {\em CVPR}, 2018.

\bibitem{sarfraz2017pose}
M.~S. Sarfraz, A.~Schumann, A.~Eberle, and R.~Stiefelhagen.
\newblock A pose-sensitive embedding for person re-identification with expanded
  cross neighborhood re-ranking.
\newblock In {\em CVPR}, 2018.

\bibitem{Schumann2017Person}
A.~Schumann and R.~Stiefelhagen.
\newblock Person re-identification by deep learning attribute-complementary
  information.
\newblock In {\em CVPR Workshops}, 2017.

\bibitem{Shen2012Object}
X.~Shen, Z.~Lin, J.~Brandt, and S.~Avidan.
\newblock Object retrieval and localization with spatially-constrained
  similarity measure and k-nn re-ranking.
\newblock In {\em CVPR}, 2012.

\bibitem{Si2018Dual}
J.~Si, H.~Zhang, C.~G. Li, J.~Kuen, X.~Kong, A.~C. Kot, and G.~Wang.
\newblock Dual attention matching network for context-aware feature sequence
  based person re-identification.
\newblock In {\em CVPR}, 2018.

\bibitem{Suh2018Part}
Y.~Suh, J.~Wang, S.~Tang, T.~Mei, and K.~M. Lee.
\newblock Part-aligned bilinear representations for person re-identification.
\newblock In {\em ECCV}, 2018.

\bibitem{Sun2017SVDNet}
Y.~Sun, L.~Zheng, W.~Deng, and S.~Wang.
\newblock Svdnet for pedestrian retrieval.
\newblock In {\em CVPR}, 2017.

\bibitem{sun2017beyond}
Y.~Sun, L.~Zheng, Y.~Yang, Q.~Tian, and S.~Wang.
\newblock Beyond part models: Person retrieval with refined part pooling.
\newblock In {\em ECCV}, 2018.

\bibitem{wang2012unsupervised}
B.~Wang, J.~Jiang, W.~Wang, Z.-H. Zhou, and Z.~Tu.
\newblock Unsupervised metric fusion by cross diffusion.
\newblock In {\em CVPR}, 2012.

\bibitem{Wang2007Shape}
X.~Wang, G.~Doretto, T.~Sebastian, J.~Rittscher, and P.~Tu.
\newblock Shape and appearance context modeling.
\newblock In {\em ICCV}, 2007.

\bibitem{Xiong2014Person}
F.~Xiong, M.~Gou, O.~Camps, and M.~Sznaier.
\newblock Person re-identification using kernel-based metric learning methods.
\newblock In {\em ECCV}, 2014.

\bibitem{Xu2018Attention}
J.~Xu, R.~Zhao, F.~Zhu, H.~Wang, and W.~Ouyang.
\newblock Attention-aware compositional network for person re-identification.
\newblock In {\em CVPR}, 2018.

\bibitem{yan2014view}
C.~Yan, S.~Shan, D.~Wang, H.~Li, and X.~Chen.
\newblock View-adaptive metric learning for multi-view person
  re-identification.
\newblock In {\em ACCV}, 2014.

\bibitem{yang2015re}
F.~Yang, B.~Matei, and L.~S. Davis.
\newblock Re-ranking by multi-feature fusion with diffusion for image
  retrieval.
\newblock In {\em WACV}, 2015.

\bibitem{yang2009locally}
X.~Yang, S.~Koknar-Tezel, and L.~J. Latecki.
\newblock Locally constrained diffusion process on locally densified distance
  spaces with applications to shape retrieval.
\newblock In {\em CVPR}, 2009.

\bibitem{yang2013affinity}
X.~Yang, L.~Prasad, and L.~J. Latecki.
\newblock Affinity learning with diffusion on tensor product graph.
\newblock {\em IEEE transactions on pattern analysis and machine intelligence},
  35(1):28--38, 2013.

\bibitem{Yang2014Salient}
Y.~Yang, J.~Yang, J.~Yan, S.~Liao, D.~Yi, and S.~Z. Li.
\newblock Salient color names for person re-identification.
\newblock In {\em ECCV}, 2014.

\bibitem{yu2013feature}
J.~Yu, Z.~Qin, T.~Wan, and X.~Zhang.
\newblock Feature integration analysis of bag-of-features model for image
  retrieval.
\newblock {\em Neurocomputing}, 120:355--364, 2013.

\bibitem{Yu2017Divide}
R.~Yu, Z.~Zhou, S.~Bai, and X.~Bai.
\newblock Divide and fuse: A re-ranking approach for person re-identification.
\newblock In {\em BMVC}, 2017.

\bibitem{zhang2012query}
S.~Zhang, M.~Yang, T.~Cour, K.~Yu, and D.~N. Metaxas.
\newblock Query specific fusion for image retrieval.
\newblock In {\em ECCV}. 2012.

\bibitem{zhang2015query}
S.~Zhang, M.~Yang, T.~Cour, K.~Yu, and D.~N. Metaxas.
\newblock Query specific rank fusion for image retrieval.
\newblock {\em IEEE Transactions on Pattern Analysis and Machine Intelligence},
  37(4):803--815, 2015.

\bibitem{zhang2017alignedreid}
X.~Zhang, H.~Luo, X.~Fan, W.~Xiang, Y.~Sun, Q.~Xiao, W.~Jiang, C.~Zhang, and
  J.~Sun.
\newblock Alignedreid: Surpassing human-level performance in person
  re-identification.
\newblock {\em arXiv preprint arXiv:1711.08184}, 2017.

\bibitem{zheng2015scalable}
L.~Zheng, L.~Shen, L.~Tian, S.~Wang, J.~Wang, and Q.~Tian.
\newblock Scalable person re-identification: A benchmark.
\newblock In {\em ICCV}, 2015.

\bibitem{zheng2015query}
L.~Zheng, S.~Wang, L.~Tian, F.~He, Z.~Liu, and Q.~Tian.
\newblock Query-adaptive late fusion for image search and person
  re-identification.
\newblock In {\em CVPR}, 2015.

\bibitem{Zheng2016Person}
L.~Zheng, H.~Zhang, S.~Sun, M.~Chandraker, Y.~Yang, and Q.~Tian.
\newblock Person re-identification in the wild.
\newblock In {\em CVPR}, 2017.

\bibitem{zheng2017pedestrian}
Z.~Zheng, L.~Zheng, and Y.~Yang.
\newblock Pedestrian alignment network for large-scale person
  re-identification.
\newblock {\em IEEE Transactions on Circuits and Systems for Video Technology},
  2018.

\bibitem{Zhong2017Re}
Z.~Zhong, L.~Zheng, D.~Cao, and S.~Li.
\newblock Re-ranking person re-identification with k-reciprocal encoding.
\newblock In {\em CVPR}, 2017.

\end{thebibliography}
}

\end{document}